\documentclass[]{bytedance_seed}

\usepackage[toc,page,header]{appendix}

\usepackage{minitoc}

\usepackage{ulem}

\usepackage[numbers]{natbib}
\usepackage{placeins}
\usepackage{makecell} 
\usepackage{graphicx}
\usepackage{fix-cm}
\usepackage{multirow}
\usepackage{enumitem}
\usepackage{amsmath}
\usepackage{bm}
\usepackage{ulem}
\usepackage{tcolorbox}
\usepackage{lipsum}
\usepackage{wrapfig}
\usepackage{booktabs}   
\usepackage{makecell}   
\usepackage{caption}    
\usepackage{tabularx}
\usepackage[utf8]{inputenc} 
\usepackage[T1]{fontenc}    
\usepackage{hyperref}       
\usepackage{url}            
\usepackage{booktabs}       
\usepackage{amsfonts}       
\usepackage{nicefrac}       
\usepackage{microtype}      
\usepackage{xcolor}         
\usepackage{amsmath}
\usepackage{xspace}
\usepackage{ulem} 
\usepackage{tabularx}
\hypersetup{
    urlcolor=blue          
}

\title{ IRASim: A Fine-Grained World Model for \\ Robot Manipulation }

\author{
  {\fontsize{11pt}{14pt}\selectfont
    Fangqi Zhu$^{1,2}$, 
    Hongtao Wu$^{2,\dagger, *}$,
    Song Guo$^{1, *}$, 
    Yuxiao Liu$^{2}$,
    Chilam Cheang$^{2}$,
    Tao Kong$^{2}$ \\
  }
  {\fontsize{10pt}{12pt}\selectfont $^{1}$Hong Kong University of Science and Technology} 
  {\fontsize{10pt}{12pt}\selectfont $^{2}$ByteDance Seed}
}

\contribution[\dagger]{Project Lead}
\contribution[*]{Corresponding Author}

\abstract{
World models allow autonomous agents to plan and explore by predicting the visual outcomes of different actions. 
However, for robot manipulation, it is challenging to accurately model the fine-grained robot-object interaction within the visual space using existing methods which overlooks precise alignment between each action and the corresponding frame.
In this paper, we present \ourmethod, a novel world model capable of generating videos with fine-grained robot-object interaction details, conditioned on historical observations and robot action trajectories.
We train a diffusion transformer and introduce a novel frame-level action-conditioning module within each transformer block to explicitly model and strengthen the action-frame alignment.
Extensive experiments show that: 
(1) the quality of the videos generated by our method surpasses all the baseline methods and scales effectively with increased model size and computation;
(2) policy evaluations using \ourmethod exhibit a strong correlation with those using the ground-truth simulator, highlighting its potential to accelerate real-world policy evaluation; 
(3) testing-time scaling through model-based planning with \ourmethod significantly enhances policy performance, as evidenced by an improvement in the IoU metric on the Push-T benchmark from 0.637 to 0.961;
(4) \ourmethod provides flexible action controllability, allowing virtual robotic arms in datasets to be controlled via a keyboard or VR controller.

}

\def\ourmethod{IRASim\xspace}
\def\projectpage{https://gen-irasim.github.io/}

\date{\today}
\checkdata[Corresponding Email]{\email{wuhongtao.123@bytedance.com}; \email{songguo@cse.ust.hk}}
\checkdata[Project Page]{\url{\projectpage}}

\begin{document}
\maketitle


\section{Introduction}
\label{sec:intro}
\begin{figure}[t] 
    \centering
    \includegraphics[width=1.0\linewidth ]
    {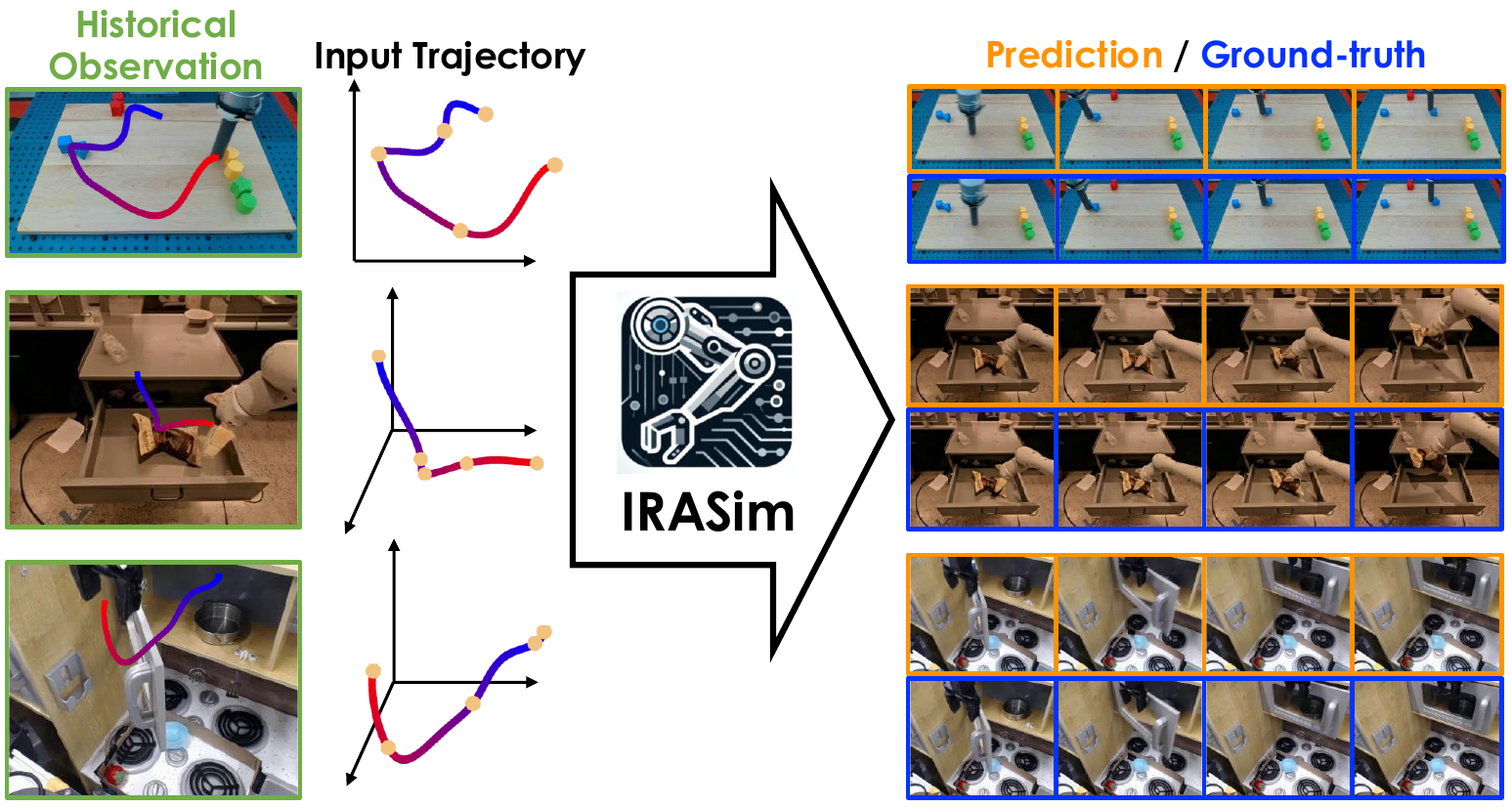}
    \caption{\textbf{Overview of \ourmethod.} \ourmethod is a fine-grained world model for robot manipulation. It generates high-fidelity videos that simulate accurate robot-object interactions of a robot executes an action trajectory, given historical observation.
    }
    \label{fig:intro}
\end{figure}


World models empower agents to foresee the outcomes of their actions by learning the fundamental dynamics of the world~\cite{agarwal2025cosmos, yang2024learning}.
This capability offers two key applications for robot manipulation.
Firstly, it allows robots to improve autonomous policies by exploring various action proposal in the model and selecting the optimal one for executing.
Secondly, world models offer the potential for scalable policy evaluation -- they can generate realistic and reasonable physical interactions, providing an efficient alternative to real-world evaluation~\cite{li24simpler}.

When training a world model for robot manipulation, accurately simulating the intricate interactions between the robot, objects, and the surrounding environment remains a substantial challenge. Manipulation tasks are inherently delicate, where even subtle variations can result in task failure. Consequently, constructing a fine-grained world model is essential for faithfully capturing these precise interactions.
Moreover, modern robotic manipulation policy leverages an action chunking technique~\cite{act, chi2023diffusionpolicy} which generates action trajectories rather than single actions to accomplish complex manipulation tasks.
In this paper, we focus on building a world model that generates videos, with fine-grained robot-object interaction details, of a robot executing an action trajectory given historical observation~(Fig.~\ref{fig:intro}). 
We refer to this task as the trajectory-to-video task.
Inspired by recent advances in text-to-video generation~\cite{videoworldsimulators2024,zheng2024opensorademocratizingefficientvideo}, we leverage generative models to capture visual details that are essential for representing the intricate dynamics of the world (\textit{e.g.}, robot-object contact and object articulation).
However, text-to-video models are trained to generate videos based on high-level textual descriptions that provide general contextual cues rather than detailed, frame-level instructions.
This is different from the trajectory-to-video task in which each action in the trajectory provides an exact description of the robot's movement in each frame of the predicted video.

To bridge this gap, we introduce \ourmethod, a new world model trained with a diffusion transformer to capture complex environment dynamics. 
We incorporate a novel frame-level action-conditioning module within each transformer block, explicitly modeling and strengthening the alignment between each action and the corresponding frame.
\ourmethod can generate high-fidelity videos to simulate fine-grained robot-object interactions, as shown in Fig.~\ref{fig:intro}.
To generate a long-horizon video that completes an entire task, \ourmethod can be rolled out in an autoregressive manner and maintain temporal consistency across each generated video clip.

We perform extensive experiments on four tasks to validate the effectiveness of the proposed method: 1) trajectory-conditioned video generation, 2) policy evaluation, 3) model-based planning, and 4) flexible action controllability.
For trajectory-conditioned video generation, we validate \ourmethod on four real-robot manipulation datasets: RT-1~\citep{brohan2022rt}, Bridge~\citep{walke2023bridgedata}, Language-Table~\citep{lynch2023interactive}, and RoboNet~\citep{robonet}.
Results show that \ourmethod can generate high-quality videos of high resolution~(up to 288$\times$512) and long horizon~(more than 150 frames).
It outperforms all the comparing baseline methods in all four datasets and is more preferable in human evaluation.
In addition, it scales effectively with increased model size and computation.
For policy evaluation, we evaluate autonomous policies in both \ourmethod and the LIBERO simulation environment~\cite{liu2023libero}.
The evaluation results from \ourmethod strongly correlate with those from the ground truth simulator, indicating great potential for scalable real-world policy evaluation.
Moreover, we leverage \ourmethod as a visual dynamics model for model-based planning in both simulation and real-world settings.
\ourmethod significantly improves the policy performance on accomplishing complex manipulation tasks in both settings by allowing the policy to explore various trajectory proposals and select the optimal one for execution.
\ourmethod improves the performance~(IoU metric) of a vanilla diffusion policy on the Push-T benchmark from 0.637 to 0.961.
More importantly, the performance improvement scales well with increased test-time computation, highlighting a promising path towards \textit{test-time scaling}~\cite{deepseekai2025deepseekr1incentivizingreasoningcapability} for robot manipulation.
Finally, we demonstrate the flexible action controllability of \ourmethod by generating videos of controlling the virtual robots in the datasets via trajectories collected with a keyboard or VR controller.
We recommend visiting the \href{\projectpage}{\textcolor{blue}{project page}} for full videos.
To summarize, the contribution of this paper is threefold:
\begin{itemize}[leftmargin=5mm]
    \item We propose \ourmethod, a novel method that is capable of generating high-quality videos with fine-grained robot-object interaction details for the trajectory-to-video task. It achieves precise action-frame alignment via a novel frame-level action-conditioning module.
    \item We perform extensive experiments on trajectory-conditioned video generation.
    Results show that \ourmethod outperforms all the comparing baseline methods in video generation and scales effectively with increased model size and computation.
    \item 
    We showcase the usefulness of \ourmethod in robot manipulation through policy evaluation and policy improvement.
    We observe a strong correlation of evaluation results between evaluating in \ourmethod and the ground-truth simulator.
    When combined with model-based planning algorithm, \ourmethod improves the policy performance on accomplishing complex manipulation tasks in both simulation and the real world.
\end{itemize}

\section{Related Work}
\paragraph{World Models.}
Learning a world model~(or dynamics model)~\citep{lecun2022path, ha2018world}, which predicts future observations based on current observations and actions, has recently become increasingly popular~\citep{bruce2024genie, parkerholder2024genie2, agarwal2025cosmos}.
In autonomous driving, world models have been used to infer future states of the environment for safe and robust driving~\cite{santana2016learning, hu2023gaia, gao2025vista}.
World models are also leveraged as a promising approach for training safe and sample-efficient reinforcement learning agents in gaming~\cite{alonso2025diffusion, valevski2024diffusion}.
In robot manipulation, prior works~\citep{fitvid,maskvit} train action-conditioned video prediction models for planning.
Recently, iVideoGPT~\citep{wu2024ivideogpt} proposes to train an autoregressive transformer for action-conditioned video prediction.
VLP~\citep{du2024video} and UniSim~\cite{yang2024learning} use languages with action information to prompt text-to-video models for generating video.
\ourmethod differs from these works in that it can generate high-resolution (up to 288$\times$512) and long-horizon (up to 150+ frames) videos given the initial observation and a robot trajectory, accurately capturing fine-grained robot-object interactions.
It showcases strong capabilities in improving policy through model-based planning and potential for scalable policy evaluation.

\paragraph{Video Models.}
Video models generate video frames either unconditionally or with conditions including classes, initial frames, texts, strokes, and/or actions~\citep{finn2016unsupervised,ma2024latte,bao2024vidu,wang2024boximator}.
Recently, diffusion models~\citep{ho2020denoising} are becoming more and more popular in video generation~\citep{ho2022video, he2023latent, videoworldsimulators2024, zheng2024opensorademocratizingefficientvideo, yang2024cogvideox}.
Sora~\citep{videoworldsimulators2024} showcases extraordinary video generation capability with Diffusion Transformers~\citep{peebles2023scalable}.
\ourmethod also leverages Diffusion Transformers as the backbone.
A relevant line of work is to control video synthesis with motions.
These methods use either user-specified strokes~\citep{yin2023dragnuwa, chen2023mcdiff}, bounding boxes~\citep{wang2024boximator}, or human poses~\citep{wang2023disco, xu2023magicanimate} as conditions.
In contrast, \ourmethod models complex 2D and 3D actions over timesteps via a novel frame-level action-conditioning module.

\paragraph{Robot Learning with World Models.}
World models hold the promise of allowing the robot to predict the effects of actions and plan solutions in complex environments~\cite{maskvit, finn2017deep, gao2024flip, zhao2024vlmpcvisionlanguagemodelpredictive, vp2}.
For policy learning, prior works combine action-conditioned video prediction with model-predictive control for robot manipulation~\cite{finn2017deep, ebert2018visual, zhao2024vlmpcvisionlanguagemodelpredictive}.
DreamerV3~\citep{hafner2023mastering} and DayDreamer~\citep{wu2023daydreamer} leverage recurrent state space model~(RSSMs)~\citep{hafner2019learning} to learn a latent representation of states by modeling a world model for reinforcement learning.
Recently, FLIP~\cite{gao2024flip} proposed generating video plans that maximizes reward by leveraging flow prediction and then performing inverse dynamics to generate actions. This differs from the model-based planning we use in that we can predict the rewards of actions by predicting future videos, thereby selecting the optimal actions for execution.
To facilitate scalable policy evaluation, recent work~\cite{li24simpler} shows a correlation between evaluation in a physical simulator and on real robots.
In contrast to using a physical simulator, our work aims to leverage powerful generative models to simulate the rollouts of policies to evaluate their quality.


\section{Methods}
\begin{figure*}[t] 
    \centering
    \includegraphics[width=1.0\linewidth]{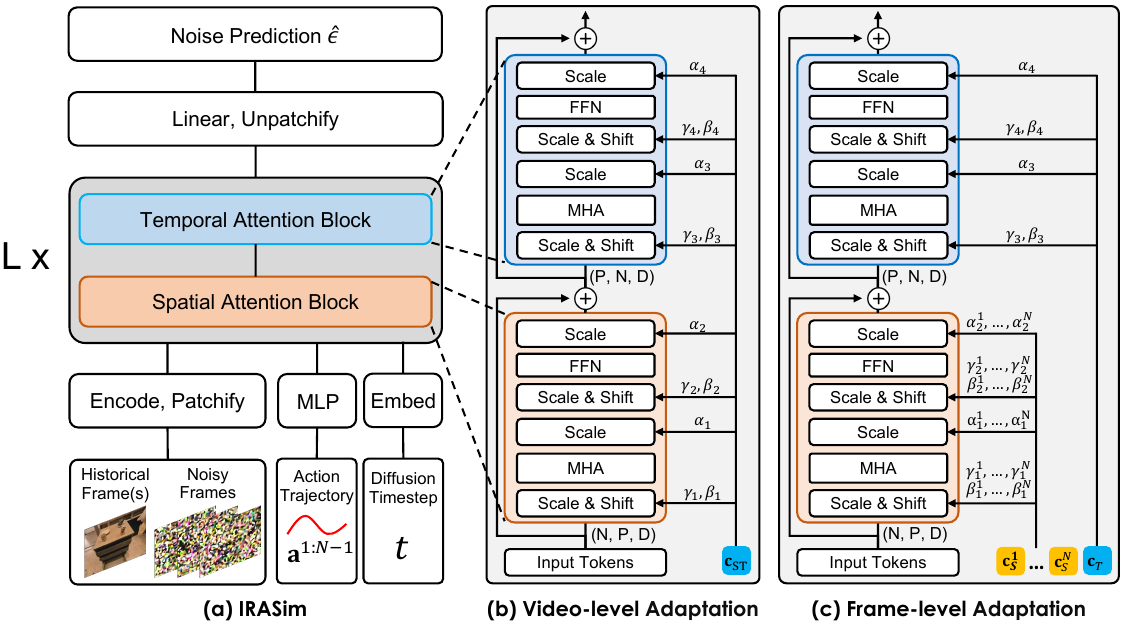}
    \caption{
    \textbf{Network Architecture of \ourmethod}. 
    (a) shows the general diffusion transformer architecture of \ourmethod. The input to \ourmethod includes the historical frames and the given trajectory.
    (b) Video-level adaptation (Video-Ada).
    (c) Frame-level adaptation (Frame-Ada).}
    \label{fig:architecture}
\end{figure*}

\subsection{Problem Statement}
\label{sec:method:problem}
We define the trajectory-to-video task as predicting the video of a robot that executes a trajectory $\mathbf{a}^{t:t+n}$ given the historical observation images $\mathbf{I}^{t-h:t}$:
\begin{equation}
\mathbf{I}^{t+1:t+n+1} = f(\mathbf{I}^{t-h:t}, \mathbf{a}^{t:t+n})
\label{eq:definition}
\end{equation}
where $h$ denotes the number of historical frames; $n$ denotes the number of actions in the video; $\mathbf{a}^{i} \in \mathbb{R}^{d}$ denotes the action at the i-th timestep.
In this paper, we focus on predicting videos for robot arms.
A typical action space of a robot arm contains 7 degrees of freedom (DoFs), \textit{i.e.}, $\mathbf{a}^{i} \in \mathbb{R}^{7}$, where 3 DoFs represent translation in the 3D space, 3 DoFs correspond to 3D rotation, and 1 DoF accounts for the gripper action. 
Additional details regarding the number of historical frames $h$ and action space dimension $d$ are provided in Appendix~\ref{app:action_space}.

\subsection{Preliminaries}
\label{sec:method:preliminaries}
Before delving into our method, we briefly review preliminaries of diffusion models~\citep{sohl2015deep, ho2020denoising}.
Diffusion models typically consist of a forward process and a reverse process.
The forward process gradually adds Gaussian noises to data $\mathbf{x}_0$ over $T$ timesteps.
It can be formulated as $q\left(\mathbf{x}_t|\mathbf{x}_{0}\right)=\mathcal{N}\left(\mathbf{x}_t; \sqrt{\overline{\alpha}_t} \mathbf{x}_{0}, 1-\overline{\alpha}_t \mathbf{I}\right)$, where $\mathbf{x}_t$ is the diffused data at the $t$-th diffusion timestep and $\overline{\alpha}_t$ is a constant defined by a variance schedule.
The reverse process starts from $\mathbf{x}_T\sim\mathcal{N}(\mathbf{0}, \mathbf{I})$ and gradually remove noises to recover $\mathbf{x}_{0}$.
It can be mathematically expressed as \( p_\theta(\mathbf{x}_{t-1}|\mathbf{x}_t) = \mathcal{N}(\mathbf{x}_{t-1}; \mu_\theta(\mathbf{x}_t, t), \Sigma_\theta(\mathbf{x}_t, t)) \), where \( \mu_\theta(\cdot) \) and \( \Sigma_\theta(\cdot) \) denote the mean and covariance functions, respectively, and can be parameterized via a neural network.

In the training phase, we sample a timestep $t \in [1, T]$ and obtain $\mathbf{x}_t = \sqrt{\overline{\alpha}_t} \mathbf{x}_0 + \sqrt{1-\overline{\alpha}_t} \epsilon_t$ via the reparameterization trick~\citep{ho2020denoising} where $\epsilon_t \in \mathcal{N}(\mathbf{0},\mathbf{I})$. 
We leverage the simplified training objective to train a noise prediction model $\epsilon_\theta$ as in DDPM~\citep{ho2020denoising}:
\begin{equation}
    \mathcal{L}_{\mathrm{simple}}(\theta) = ||\epsilon_\theta(\mathbf{x}_t,t) - \epsilon_t||^{2}
    \label{eq:noise_prediction_loss}
\end{equation}
In the inference phase, we generate $\mathbf{x}_{0}$ by first sampling $\mathbf{x}_T$ from $\mathcal{N}(\mathbf{0}, \mathbf{I})$ and iteratively compute 
\begin{equation}
 \mathbf{x}_{t-1} = \frac{1}{\sqrt{\alpha_t}} \left( \mathbf{x}_t - \frac{1 - \alpha_t}{\sqrt{1 - \bar{\alpha}_t}} \epsilon_\theta(\mathbf{x}_t, t) \right)
 \label{eq:denoise}
\end{equation}
until $t=0$.
For conditional diffusion processes, the noise prediction model $\epsilon_\theta$ can be parameterized as $\epsilon_\theta(\mathbf{x}_t,t,\mathbf{c})$ where $\mathbf{c}$ is the condition that controls the generation process.
Throughout the paper, we use superscript and subscript to indicate the timestep of a frame in the input video and the diffusion timestep, respectively.

However, directly diffusing the entire video in the pixel space is time-consuming and requires substantial computation to generate long videos with high resolutions~\citep{ho2022video}. 
Inspired by~\citet{ma2024latte}, we perform the diffusion process in a low-dimension latent space $\mathbf{z}$ instead of the pixel space for computation efficiency.
Following~\citet{he2023latent}, we leverage the pre-trained variational autoencoder (VAE) in SDXL~\citep{podell2023sdxl} to compress each frame $\mathbf{I}^{t}$ in the video to a latent representation with the VAE encoder $\mathbf{z}^{t} = \mathrm{Enc}(\mathbf{I}^{t})$.
The latent representation can be decoded back to the pixel space with the VAE decoder $\mathbf{I}^{t} = \mathrm{Dec}(\mathbf{z}^{t})$.

\subsection{IRASim}
\label{sec:irasim}

\ourmethod is a conditional diffusion model operating in the latent space of the VAE introduced in Sec.~\ref{sec:method:preliminaries}.
The condition $\mathbf{c}$ consists of the latent representation of the historical frames, $\mathbf{z}^{t-h:t} = \mathrm{Enc}(\mathbf{I}^{t-h:t})$, and an action trajectory, $\mathbf{a}^{t:t+n}$.
The diffusion target is the latent representations of the subsequent $n$ frames of the video in which the robot executes the action trajectory, \textit{i.e.} $\mathbf{x} = \mathbf{z}^{t+1:t+n+1}$. 
Inspired by Sora's remarkable capability of understanding the physical world~\citep{videoworldsimulators2024}, we similarly adopt Diffusion Transformers (DiT)~\citep{peebles2023scalable} as the backbone of \ourmethod.
In the design of \ourmethod, we aim to address three key aspects:
1) consistency with the given historical frames
2) adherence to the given action trajectory and
3) computation efficiency.
In the following, we describe pivotal design choices to achieve the aforementioned objectives.


Standard transformer blocks apply Multi-Head Self-Attention (MHA) to all tokens in the input token sequence, resulting in quadratic computation cost.
We thus leverage the memory-efficient spatial-temporal attention mechanism~\citep{xu2020spatial, bruce2024genie, ma2024latte} in the transformer blocks of \ourmethod to reduce the computation cost (Fig.~\ref{fig:architecture}).
The historical frame condition is achieved by treating the historical frames as the ground-truth portion in the input video sequence~\citep{videoworldsimulators2024}.
That is, during training, we only add noise to the tokens corresponding to the predicted frames $\mathbf{z}^{t+1:t+n+1}$, while keeping those of the historical frame $\mathbf{z}^{t-h:t}$ intact as it does not need to be predicted (Fig.~\ref{fig:architecture}).
And the diffusion loss is only computed upon the predicted frames.
This condition approach ensures consistency with the historical frames by enabling the predicted frames to interact with them via attention mechanism.

To inject the trajectory condition into video generation, we follow Diffusion Transformers~\cite{peebles2023scalable} and utilize adaptive layer normalization for conditioning.
Below, we outline two methods for incorporating the trajectory condition.
\begin{itemize}
\item \textit{Video-Level Condition.}
Similar to using a text embedding to condition the generation of the entire video in the text-to-video task, we use a linear layer to encode the trajectory into a single embedding for condition.
The embedding is then added to the embedding of the diffusion timestep $t$ for generating the scale parameters $\gamma$ and $\alpha$ and the shift parameters $\beta$ for each spatial and temporal attention block.
The overall framework is shown in Figure~\ref{fig:architecture}(b).
See Appendix~\ref{app:model_details_video} for more details.
\item \textit{Frame-Level Condition.}
Unlike the text-to-video task where the text describes the entire video, the trajectory in the trajectory-to-video task is a finer description.
Each action in the trajectory defines how the robot should move in each frame.
And thus, each generated frame must match with its corresponding action in the trajectory.
To achieve this precise frame-level alignment, we condition the generation of each frame by its corresponding action.
Instead of encoding the action trajectory into a single embedding, we use a linear layer to encode each action into an individual embedding.
The diffusion timestep embedding is added to each action embedding to generate the scale and shift parameters for each individual frame in the spatial block.
The scale and shift parameters of the temporal block for all frames share the same conditioning embedding which is derived similarly as in video-level condition.
The overall framework is shown in Figure~\ref{fig:architecture}(c).
See Appendix~\ref{app:model_details_frame} for more details.
\end{itemize}

The output layer contains a linear layer which outputs the noise prediction $\hat{\epsilon} = \epsilon_{\theta}(\mathbf{x}_{t}, t, \mathbf{c})$.
$\hat{\epsilon}$ is used to compute the L2 loss with the ground-truth noise during training (Eq.~\ref{eq:noise_prediction_loss}).
Note that the VAE is frozen during the whole training process.
During inference, we sample $\mathbf{x}^{T}$ from $\mathcal{N}(\mathbf{0}, \mathbf{I})$ and gradually denoise it via Eq.~\ref{eq:denoise} to obtain the  latent representation of the predicted frames $\hat{\mathbf{z}}^{t+1:t+n+1} = \mathbf{x}_{0}$.
The predicted frames can be decoded with the VAE decoder $\hat{\mathbf{I}}^{t+1:t+n+1}=\mathrm{Dec}(\hat{\mathbf{z}}^{t+1:t+n+1})$.

\section{Experiments}

\begin{figure*}[htbp] 
    \centering
    \includegraphics[width=1.0\linewidth]{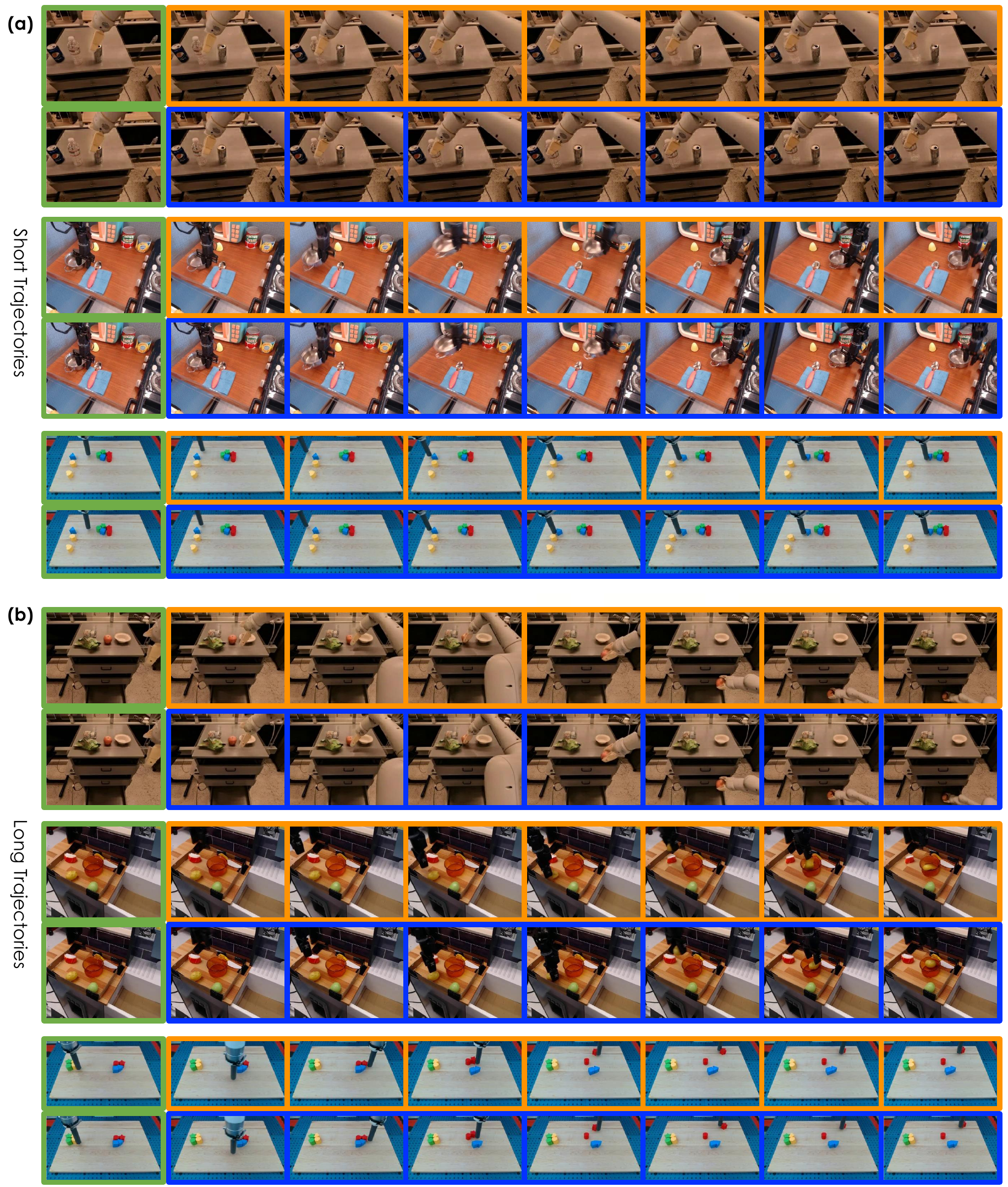}
    \caption{
        \textbf{Qualitative Results.}
        We show video generation of \ourmethod with (a) short trajectories and (b) long trajectories on the test set of RT-1, Bridge, and Language-Table. Ground-truths are in blue boxes. Predictions are in orange boxes. Initial ground-truth video frames are in green boxes. Please see our \href{\projectpage}{\textcolor{blue}{project page}} for videos.
    }
    \label{fig:short_long}
\end{figure*}

We perform extensive experiments to validate the effectiveness of \ourmethod. 
We aim to answer three main questions: 
1) Is \ourmethod effective on modeling fine-grained robot-object interactions and solving the trajectory-to-video task on various real-robot datasets with different action spaces?
2) Can we leverage \ourmethod as a world model for policy evaluation on manipulation tasks?
3) Can we utilize \ourmethod for model-based planning and improve flat autonomous policies on manipulation tasks?
We also perform extensive ablation studies to analyze the contribution of different components of the proposed method.

\subsection{Trajectory-Conditioned Video Prediction}
\paragraph{Experiment Setup} \label{exp:setup} We conduct experiments on four real-robot manipulation datasets: RT-1~\citep{brohan2022rt}, Bridge~\citep{walke2023bridgedata}, Language-Table~\citep{lynch2023interactive}, and RoboNet~\citep{robonet}. 
The action space varies across datasets, with RT-1 and Bridge using 7 DoF, Language-Table 2 DoF, and RoboNet up to 5 DoF. 
Details of each dataset are provided in Appendix~\ref{app:dataset}.
For RT-1, Bridge, and Language-Table, we use 1 historical frame and 15 actions as context to predict the next 15 frames.
For RoboNet, we follow iVideoGPT~\cite{wu2024ivideogpt} and use 2 historical frames and 10 actions to predict the next 10 frames. 
Videos are resized to 256$\times$320 for RT-1 and Bridge, 288$\times$512 for Language-Table, and 256$\times$256 for RoboNet.
We evaluate video generation on both short and long trajectories.
Short trajectories contains $n$ actions and the videos can be generated in a single generation pass.
Long trajectories consists of more actions and the videos are generated autoregressively over multiple passes.
The final generated frame from the previous pass serves as the conditional historical frame for the current one.
We denote video-level and frame-level adaptation as \ourmethod-Video-Ada and \ourmethod-Frame-Ada, respectively, and refer to them as Frame-Ada and Video-Ada for brevity.
Training details can be found in Appendix~\ref{app:training_details}.

\paragraph{Baselines.}
We compare \ourmethod with two state-of-the-art methods, \textit{i.e.}, VDM~\citep{ho2022video} and LVDM~\citep{he2023latent}. 
Both methods are diffusion models based on a U-Net architecture.
This is in contrast to \ourmethod, which employs a Transformer architecture. LVDM diffuses videos in a latent space, while VDM operates in the pixel space. 
To impose trajectory conditions on video generation, we encode the trajectory into an embedding to condition the diffusion process for both methods. 
This is similar to the text embedding used for text-to-video generation in the original papers~\citep{ho2022video,he2023latent}. 
Additionally, we compare with two state-of-the-art non-diffusion methods, iVideoGPT~\citep{wu2024ivideogpt} and MaskViT~\citep{maskvit}, on the RoboNet dataset.
iVideoGPT autoregressively predicts the next visual token; MaskVit generates visual tokens via a iterative refinement process.
More details about baselines can be found in Appendix~\ref{app:baselines}.

\begin{table}[t]
\centering
\begin{tabular}{c|c|cc|ccc}
\toprule
\multirow{2}{*}{Dataset} & \multirow{2}{*}{Method} & \multicolumn{2}{c|}{Computation-based} & \multicolumn{3}{c}{Model-based} \\ \cmidrule(r){3-4} \cmidrule(r){5-7}
 &  & \textbf{PSNR} $\uparrow$ & SSIM $\uparrow$ & \textbf{Latent L2} $\downarrow$ & FID $\downarrow$ & FVD $\downarrow$ \\ 
\midrule
\multirow{4}{*}{RT-1} & VDM & 13.762 & 0.554 & 0.4983 & 41.23 & 371.13\\ 
                     & LVDM & 25.041 & 0.815 & 0.2244 & \textbf{4.26} & 30.72\\  
                     & Video-Ada & \underline{25.446} & \underline{0.823} & \underline{0.2191} & \underline{4.34} & \underline{29.27} \\  
                     & \textbf{Frame-Ada (Ours)} & \textbf{26.048} & \textbf{0.833} & \textbf{0.2099} & 5.60 & \textbf{25.58} \\ \midrule
\multirow{4}{*}{Bridge} & VDM & 18.520 & 0.741 & 0.3709 & 39.82 & 127.25\\
                        & LVDM & 23.546 & 0.810 & 0.2155 & 10.59 & 35.06\\ 
                        & Video-Ada & \underline{24.733} & \underline{0.827} & \underline{0.2021} & \textbf{10.30} &  \underline{23.03} \\  
                        & \textbf{Frame-Ada (Ours)} & \textbf{25.275} & \textbf{0.833} & \textbf{0.1947} & \underline{10.51} & \textbf{20.91} \\ \midrule 
\multirow{4}{*}{\shortstack{Language \\ Table}} & VDM & 23.067 & 0.857 & 0.3204 & 64.63 & 136.56 \\  
                     & LVDM & \underline{28.254} & \textbf{0.889} & \underline{0.1704} & \underline{6.85} & \textbf{24.34} \\  
                     & Video-Ada & 23.893 & 0.859 & 0.2028 & 7.05 & 73.84\\  
                     & \textbf{Frame-Ada (Ours)} & \textbf{28.818} & \underline{0.888} & \textbf{0.1660} & \textbf{6.38} & \underline{48.49} \\ \bottomrule              
\end{tabular}
\caption{\textbf{Quantitative results for short-trajectory video generation.} We prioritize Latent L2 and PSNR as the primary evaluation metrics. Video-Ada and Frame-Ada are variants of \ourmethod.}
\label{tab:short_video}
\end{table}

\begin{table}[!htbp]
\centering
\begin{tabular}{c|c|c}
\toprule
Method & PSNR $\uparrow$ & SSIM $\uparrow$ \\
\midrule
MaskViT~\cite{maskvit}*  & 20.4 & 67.1 \\
iVideoGPT~\cite{wu2024ivideogpt}*  & 23.8 & 80.8 \\
\textbf{IRASim (Ours)} & \textbf{24.6} & \textbf{81.1} \\
\bottomrule
\end{tabular}
\caption{\textbf{Quantitative results for video generation on RoboNet dataset.} * indicates that the result is derived from~\cite{wu2024ivideogpt}.}
\label{tab:robonet}
\end{table}

\begin{table}[b]
\centering
\begin{tabular}{ccccccc}
\toprule
     & \multicolumn{2}{c}{RT-1}  & \multicolumn{2}{c}{Bridge} & \multicolumn{2}{c}{Language-Table} \\
\cmidrule(r){2-3} \cmidrule(l){4-5} \cmidrule(l){6-7}
     & Latent L2 $\downarrow$ & PSNR $\uparrow$ & Latent L2 $\downarrow$ & PSNR $\uparrow$ & Latent L2 $\downarrow$ & PSNR $\uparrow$ \\
\midrule
LVDM~\cite{he2023latent} & 0.2567 & 23.573 & 0.2534 & 21.792 & \underline{0.1776} & \underline{26.215} \\
\ourmethod-Video-Ada & \underline{0.2519} & \underline{23.984} & \underline{0.2385} & \underline{22.868} & 0.2112 & 22.551 \\
\textbf{\ourmethod-Frame-Ada} & \textbf{0.2408} & \textbf{24.615} & \textbf{0.2306} & \textbf{23.260} & \textbf{0.1730} & \textbf{26.773} \\ 
\bottomrule
\end{tabular}
\caption{\textbf{Quantitative Results on Video Generation of Long Trajectories.}}
\label{tab:long_video}
\end{table}

\begin{figure*}[htbp]
    \centering
    \includegraphics[width=1.0\linewidth]{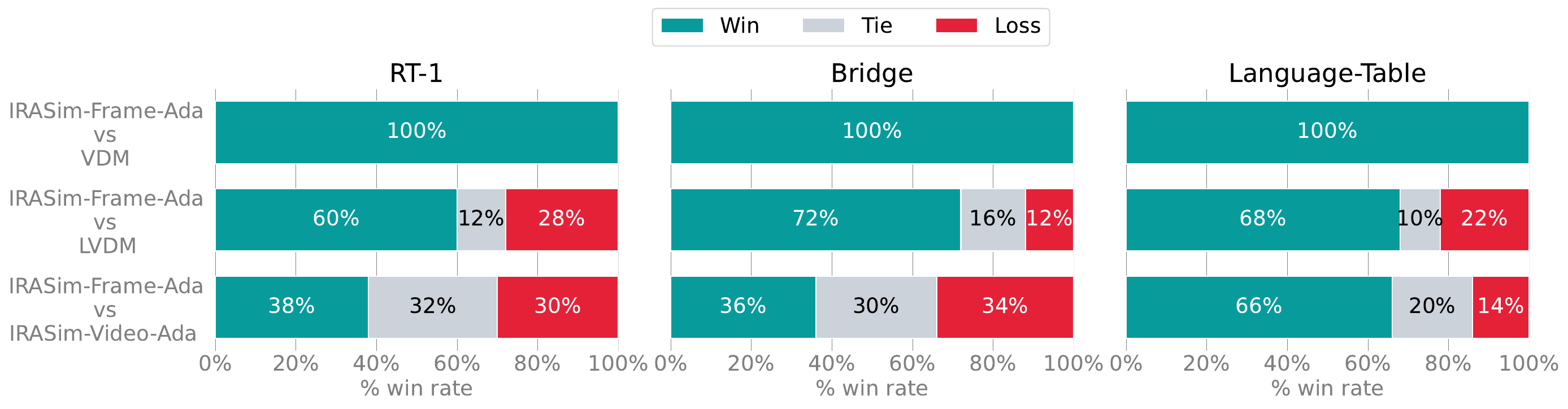}
    \caption{\textbf{Human Preference Evaluation.} We perform a user study to evaluate the human preference between \ourmethod-Frame-Ada and other baseline methods.
    }
    \label{fig:human}
\end{figure*}

\paragraph{Metrics.}  
Following~\citep{xu2023magicanimate}, we evaluate the performance with two types of metrics: computation-based (PSNR~\citep{5596999}, SSIM~\citep{1284395}) and model-based (Latent L2 loss, FID~\citep{fid}, and FVD~\citep{fvd}).  
Unlike the text-to-video task, where various videos may satisfy a single text condition, the trajectory-to-video task has much less variation: the robot in the predicted video must strictly follow the input trajectory.  
Therefore, we use video reconstruction metrics, Latent L2 loss and PSNR, as the primary evaluation metrics. 
In Appendix~\ref{app:human}, we showcase that Latent L2 loss and PSNR best align with human preferences among all the evaluated metrics. 
More details about evaluation can be found in Appendix~\ref{app:evaluate}.

\begin{figure}[t] 
    \centering
    \includegraphics[width=1.0\linewidth]{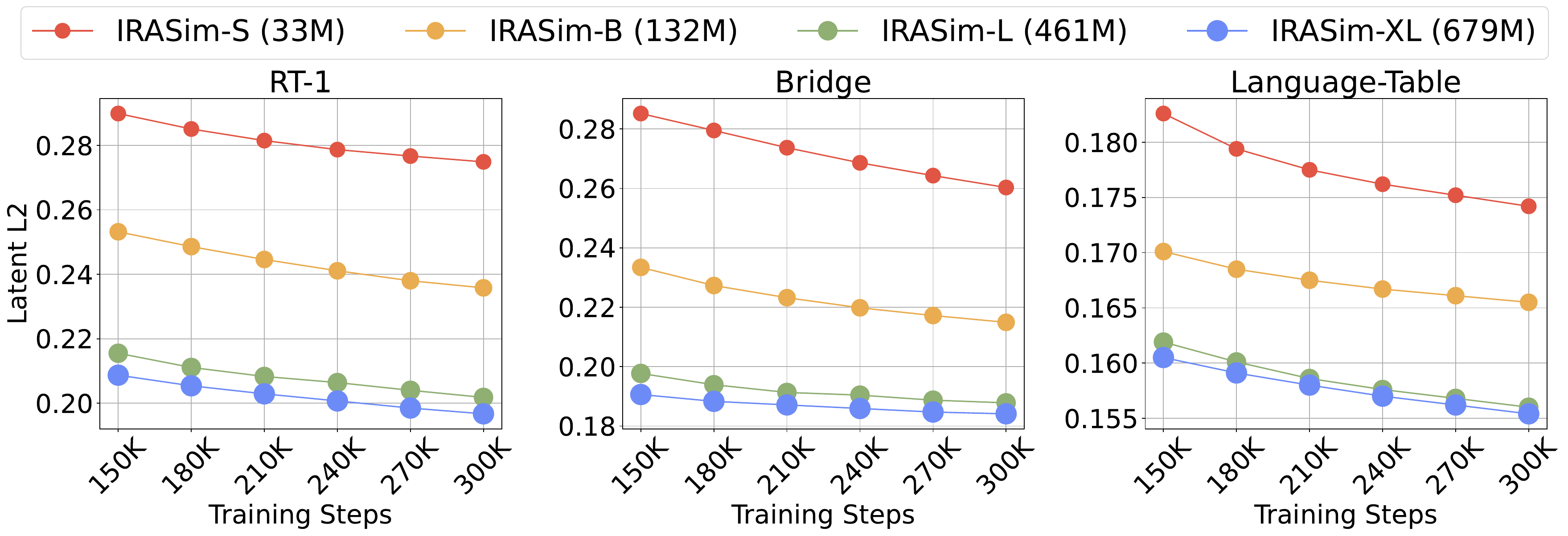}
    \caption{
        \textbf{Scaling.} \ourmethod scales effectively with the increase of model sizes and training steps.
    }
    \label{fig:scale}
\end{figure}

\paragraph{Video Generation of Short Trajectories.}
Qualitative results are shown in Fig.~\ref{fig:short_long}(a) and Fig.~\ref{fig:short_compare}.
Quantitative results are shown in Tab.~\ref{tab:short_video}.
As shown in Fig.~\ref{fig:short_long}(a) and Fig.~\ref{fig:short_compare}, \ourmethod-Frame-Ada effectively models fine-grained robot-object interactions
and generates high-quality videos that closely align with the ground truth.
It surpasses all the comparing baseline methods in our primary evaluation metrics, Latent L2 and PSNR, as well as the human evaluation in Sec~\ref{para:human}. 
As illustrated in Appendix~\ref{app:short_compare} \& A~\ref{fig:long_compare}, baseline methods struggle to guide the robot arm along the given trajectory and fail to realistically simulate interactions between the robot and the objects.

\paragraph{Video Generation of Long Trajectories.}
Qualitative results are shown in Fig.~\ref{fig:short_long}(b) and Fig.~\ref{fig:long_compare}.
Quantitative results are shown in Tab.~\ref{tab:long_video}.
We compare \ourmethod with the best baseline method LVDM~\cite{he2023latent}.
\ourmethod-Frame-Ada consistently outperforms the comparison methods in all three datasets on Latent L2 loss.
Fig.~\ref{fig:short_long}(b) and Fig.~\ref{fig:long_compare} show that it retains the powerful capability of generating visually realistic and accurate videos as in the short trajectory setting. 

\paragraph{Human Preference Evaluation.}
\label{para:human}
We also perform a user study to help understand human preferences between \ourmethod-Frame-Ada and other methods.
We juxtapose the videos of predicted by \ourmethod-Frame-Ada and the comparing method and ask humans which one they prefer.
The ground-truth is also provided as a reference.
\ourmethod-Frame-Ada beats all the comparing methods in all three datasets (Fig.~\ref{fig:human}).
This result aligns with the Latent L2 loss which justifies the reason for using Latent L2 loss as one of the primary evaluation metrics.

\paragraph{Scaling.}
We follow~\cite{peebles2023scalable} and train \ourmethod-Frame-Ada with different model sizes, ranging from 33M to 679M. 
Detailed parameters of these models are shown Appendix~\ref{app:training_details}.
Results are shown in Fig.~\ref{fig:scale}.
Across all three test datasets, \ourmethod scales effectively with larger model size and more training steps, highlighting its strong potential for further performance gains through increased computation.

\begin{figure}[htbp] 
    \centering
    \includegraphics[width=1.0\linewidth]{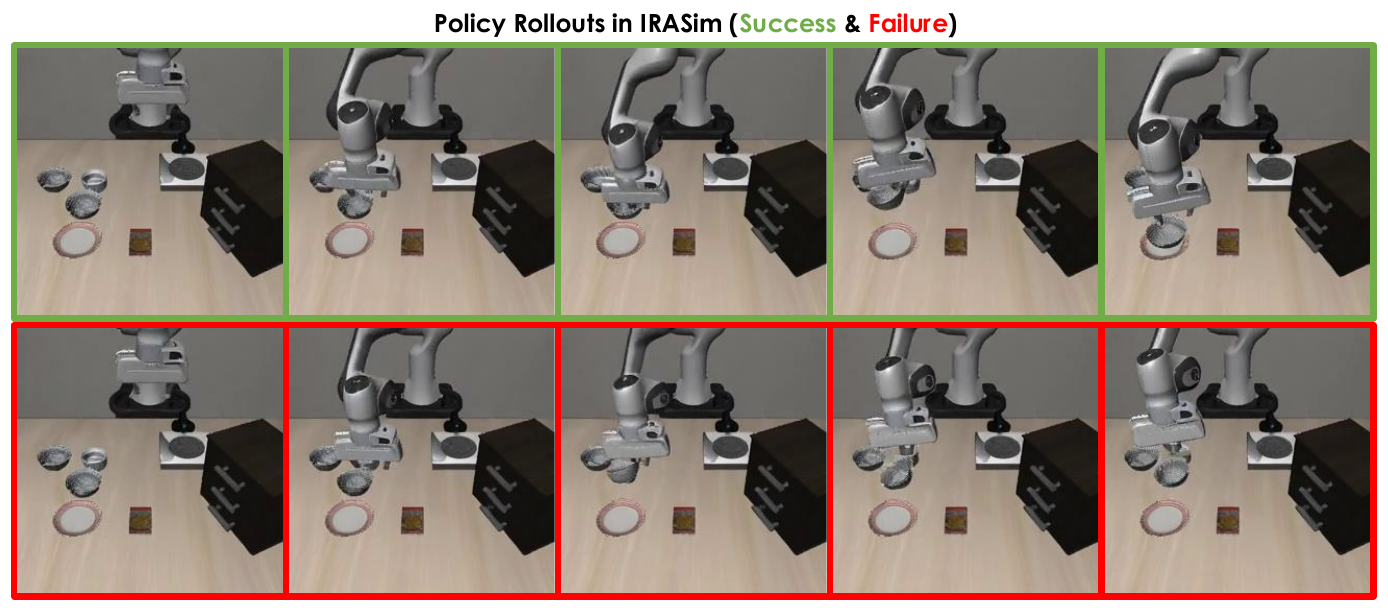}
    \caption{
    \textbf{Policy Evaluation with \ourmethod.}
    \ourmethod can simulate both successful and failed rollouts. 
    Notably, it is able to simulate a bowl slipping from the gripper.
    }
    
    \label{fig:policy_eval}
\end{figure}

\begin{table}[htbp]
    \centering
    \begin{tabular}{lcccc}
        \toprule
        Evaluator & 1 & 2 & 3 & 4 \\
        \midrule
        Ground-Truth Simulator & 0.18 & 0.50 & 0.80 & 1.00 \\
        \textbf{IRASim (Ours)} & 0.28 & 0.48 & 0.74 & 0.96 \\
        \bottomrule
    \end{tabular}
    \caption{Success rates of four different models evaluated in the two evaluators. We observe a strong correlation between the evaluation results from the ground-truth Mujoco simulator and \textbf{IRASim}. The Pearson correlation coefficient between the two evaluations is \textbf{0.99}.}
    \label{tab:policy_eval}
\end{table}

\subsection{Policy Evaluation}
\label{sec:exp:policy_evaluation}
In this section, we showcase that we can use \ourmethod as a simulator for policy evaluation.
We use the LIBERO simulation benchmark~\cite{liu2023libero} as a controlled environment for this experiment.
In particular, we evaluate a diffusion policy~\cite{chi2023diffusionpolicy, dong2024cleandiffuser} in \ourmethod and compare the evaluation results against those with the ground-truth simulator.
We train the diffusion policy on expert trajectories provided by the benchmark.
An evaluator must be able to simulate both successful and failed rollouts.
And the world model needs to learn from a broader set of data than the expert demonstrations, which contain only successful rollouts, in order to simulate both successes and failures accurately.
Thus, we deploy the trained policy in the simulator to gather additional rollouts which contains both successes and failures.
We refer to these rollouts as \textit{post-trained rollouts}.
The post-trained rollouts, along with the expert demonstrations, are used for training \ourmethod.
Given the limited amount of training data, we initialize \ourmethod with the pre-trained weight of OpenSora~\cite{zheng2024opensorademocratizingefficientvideo} to expedite the training process. 
We incorporate our frame-level condition method (Sec.~\ref{sec:irasim}) to inject the trajectory condition into the model for trajectory-conditioned video generation.

We train the diffusion policy with four different steps on the task of "pick up the black bowl between the plate and the ramekin and place it on the plate", resulting in four different individual models.
We then evaluate the performance of these four models in both the Mujoco simulator of the LIBERO benchmark and \ourmethod.
The Mujoco simulator serves as a ground truth for comparison.
We evaluate each model in both \ourmethod and the ground-truth simulator for 50 runs each.
The rollouts generated by \ourmethod were assessed by humans to determine their success or failure.

Fig.~\ref{fig:policy_eval} shows successful and failed rollouts generated by \ourmethod. 
Notably, \ourmethod successfully simulates scenarios where the bowl slips from the gripper, demonstrating strong capabilities to model fine-grained robot-object interaction.
Tab.~\ref{tab:policy_eval} reports the success rates of different models evaluated with the ground-truth Mujoco simulator and \ourmethod. 
The Pearson correlation coefficient between the two evaluation results is 0.99, indicating a strong correlation between evaluating in the ground-truth Mujoco simulator and \ourmethod.
This result showcases the potential of levaraging \ourmethod as a world model for scalable real-world policy evaluation.

\begin{figure}[htbp] 
    \centering
    \includegraphics[width=1.0\linewidth]{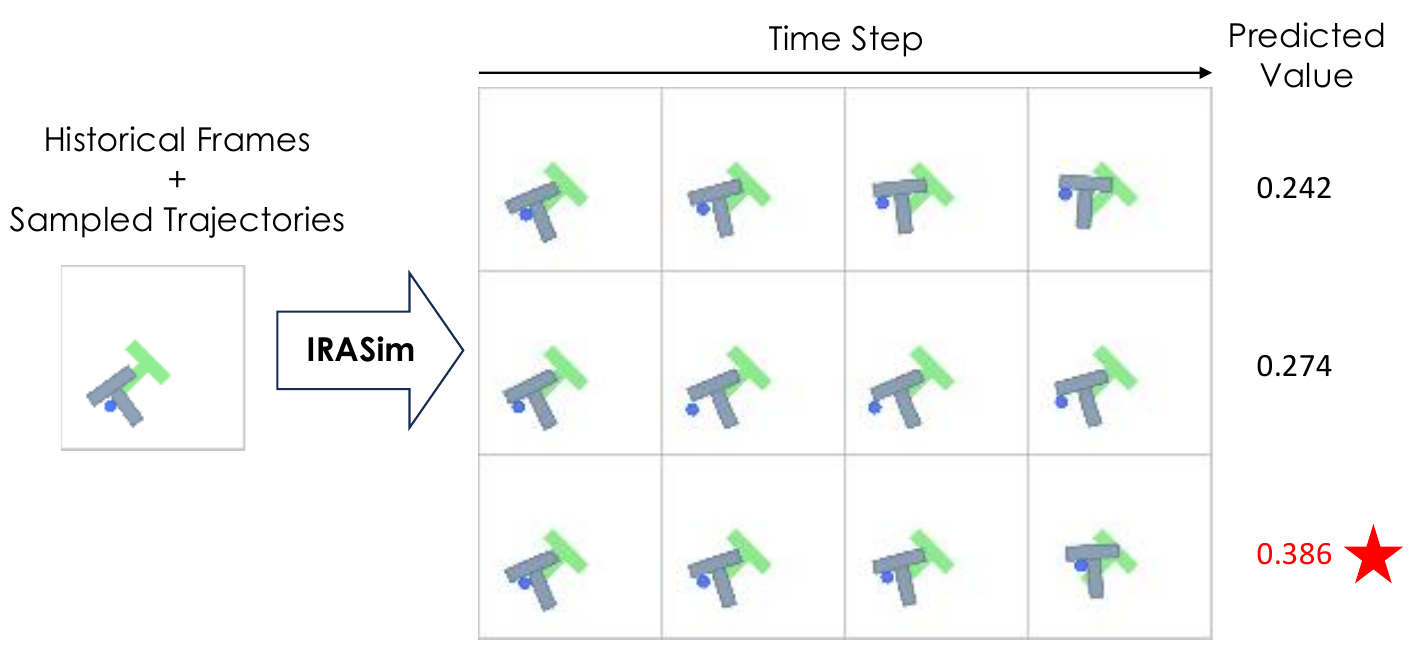}
    \caption{
        \ourmethod simulates the visual outcomes of different trajectories sampled from the policy and predicts the value of the final frame with a reward model. 
        By selecting and executing the trajectory with the highest predicted value, we enhance the existing policy by spending more time thinking (test-time compute).
    }
    
    \label{fig:pusht}
\end{figure}

\subsection{Model-based Planning for Policy Improvement}
In this section, we perform experiments in both simulation and real-world settings to show that \ourmethod can be used as a world model for model-based planning to improve vanilla policies without planning on accomplishing complex manipulation tasks.
Specifically, we adapt a simple ranking algorithm for model-based planning: 1) samples $K$ trajectories from the policy, 2) unroll each trajectory in \ourmethod, and 3) select the trajectory with the highest value for executing.

\paragraph{Push-T Simulation.}
In this experiment, we use the Push-T simulation benchmark from~\cite{florence2022implicit} for evaluation.
The robot is tasked to push a T-shaped block (gray) to a target (green) with a circular end-effector (blue) (Fig.~\ref{fig:pusht}).
In order to perform effective model-based planning, a challenge is that the world model need to accruately predict the complex dynamics of robot-block contact.
We first train a diffusion policy with 200 expert demonstrations.
Similar to Sec.~\ref{sec:exp:policy_evaluation}, we then collect post-trained rollouts, which contains both successful and failed rollouts, with the trained policy.
We use intersection over union (IoU) between the block and the target as the value function for model-based planning.
To predict the IoU of a given observation, we train a ResNet50 model~\cite{resnet} using the post-trained rollouts. 
Similar to the experiments in Sec.~\ref{sec:exp:policy_evaluation}, we initialize \ourmethod with the pre-trained weights of OpenSora and train it on both post-trained rollouts and expert demonstrations. 
We perform ablation studies to analyze the effect of varying the number of post-trained rollouts (denoted as \(P\)) on overall performance.  

\begin{table}[hbp]
    \centering
    \begin{tabular}{c c c c c c c c c c c}
        \toprule
        \centering & $P$ & $K=1$ & $K=5$ & $K=10$ & $K=50$ \\
        \midrule
        GPC-RANK & N/A* & 0.642  & - & - & 0.698 \\
        GPC-RANK+OPT & N/A* & 0.642  & 0.824 & 0.882 & - \\
        \midrule
        \multirow{5}{*}{\textbf{\ourmethod (Ours)}}
        & 0   & 0.637  & 0.679 & 0.572 & 0.418 \\
        & 100 & 0.637  & 0.847 &0.878 & 0.888 \\
        & 200 & 0.637  & 0.866 & 0.916 &0.912 \\
        & 500 & 0.637  & \textbf{0.907} & 0.906 & 0.938 \\
        & 1000 & 0.637 & 0.886 & \textbf{0.945} & \textbf{0.961} \\
        \bottomrule
    \end{tabular}
    \caption{
        \textbf{Results on Push-T Benchmark.}
        \( K \) denotes the number of sampled trajectory.
        \( P \) denotes the number of post-trained rollouts used for training \ourmethod.
        We report the average IoU over 100 trials.
        *GPC also uses additional rollouts beyond expert demonstrations to train the world model, but the number of these rollouts is not available in the paper~\cite{qi2025strengtheninggenerativerobotpolicies}.
    }
    \label{tab:pusht}
\end{table}

We compare with a recent state-of-the-art method, generative predictive control (GPC)~\cite{qi2025strengtheninggenerativerobotpolicies}.
GPC perform autoregressive next-frame prediction via diffusion to generate a video.
This contrasts with our trajectory-to-video approach, which generates all frames for a trajectory simultaneously.
Similar to \ourmethod, GPC also enhances its video prediction with additional rollouts beyond expert demonstrations.
And it also uses a diffusion policy to generate action proposals.
Specifically, we compare with two variants of GPC introduced in~\cite{qi2025strengtheninggenerativerobotpolicies}, \textit{i.e.}, GPC-RANK and GPC-RANK+OPT.
GPC-RANK uses a similar ranking-based planning algorithm as our method.
GPC-RANK+OPT utilizes a differentiable reward model to optimize action proposals via gradient optimization. 
In Tab.~\ref{tab:pusht}, \(M\) denotes the number of gradient optimization steps, and GPC-RANK+OPT represents the approach that incorporates both the RANK method and gradient optimization.
\begin{figure}[!tp] 
    \centering
    \includegraphics[width=1.0\linewidth]{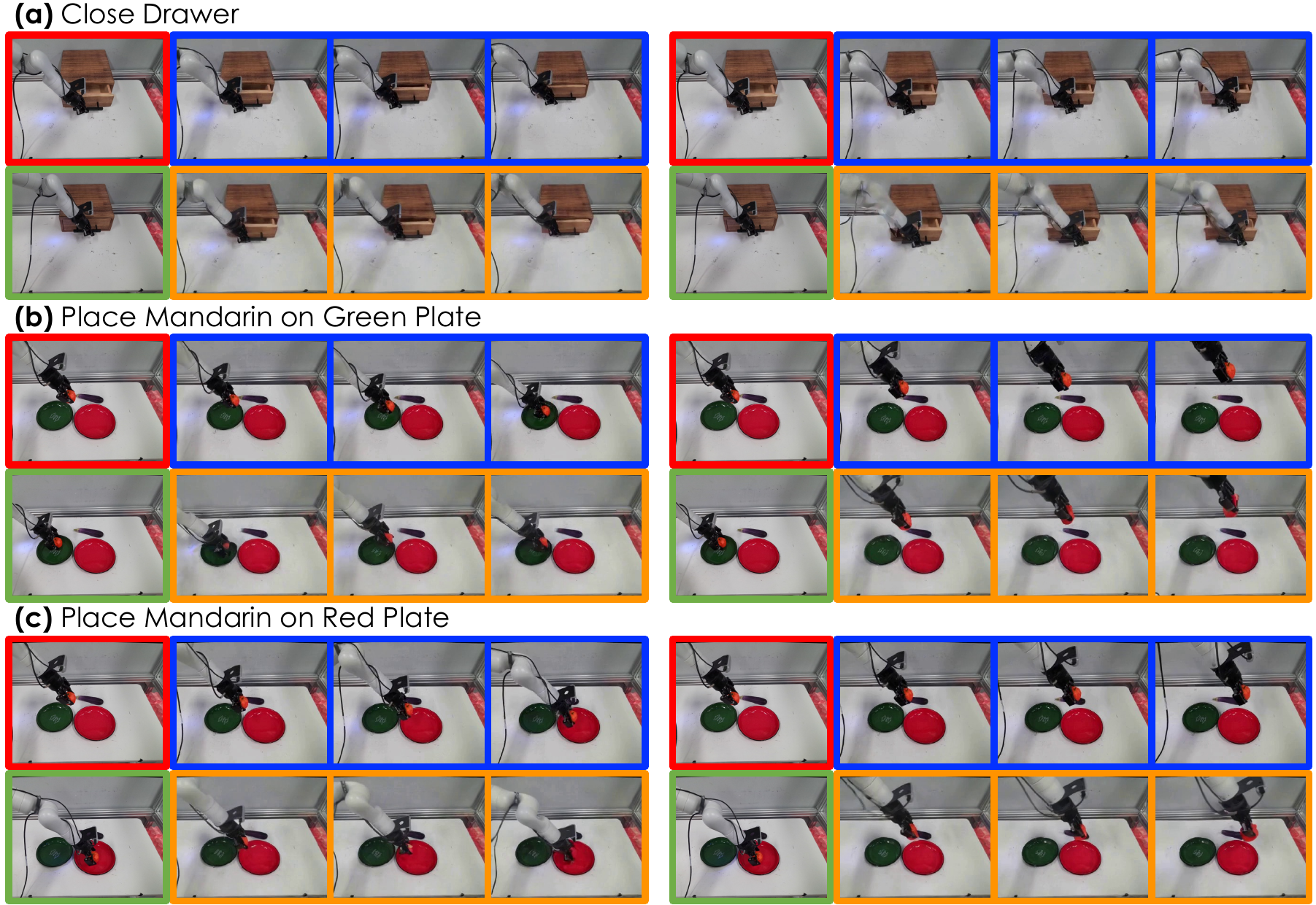}
    \caption{
        \textbf{Qualitative Results on Real-Robot Model-based Planning.}
        Historical frames are highlighted in red boxes, goal images in green boxes, real-robot rollouts in blue boxes, and videos generated by \ourmethod are shown in orange boxes.
        \ourmethod can generate videos, that faithfully matches with the ground truth in tasks involving object transportation and articulated object manipulation, enabling effective model-based planning.
    }
    \label{fig:mpc}
\end{figure}

\begin{table}[!bp]
    \setlength{\tabcolsep}{3.5pt}
    \centering
    \begin{tabular}{l c c c}
        \toprule
        \multirow{2}{*}{Method} & Close & Place Mandarin & Place Mandarin \\
        & Drawer & on Green Plate & on Red Plate \\
        \midrule
        Random & 0.20 & 0.07 & 0.13 \\
        \ourmethod~(ResNet50) & 0.60 & 0.73 & 0.60 \\
        \ourmethod~(MSE) & \textbf{0.87} & \textbf{0.80} & \textbf{0.87} \\
        \bottomrule
    \end{tabular}
    \caption{\textbf{Quantitative Results on Real-Robot Model-based Planning.}}
    \label{tab:mpc}
\end{table}

Results are shown in Tab.~\ref{tab:pusht}.
The $K=1$ column show the performance of the vanilla diffusion policy without model-based planning.
To ensure a fair comparison, we train our diffusion policy such that its IoU performance matches with that of reported in the GPC paper~\cite{qi2025strengtheninggenerativerobotpolicies}, \textit{i.e.}, 0.637 v.s. 0.642.
Using 200 post-trained rollouts, \ourmethod outperforms the two GPC variants.
And the advantage grows as more post-trained rollouts are used.
In addition, when $K=50, P=1000$, \ourmethod improves the IoU of the vanilla policy from 0.637 to 0.961.
We further explore the effect of varying \(K\) and \(P\).
When $P>0$, the policy performance consistently improves as the number of sampled trajectories $K$ increases.
This highlights the importance of including post-trained rollouts in training the world model.
More importantly, this result indicates we can robustly improve policy performance by scaling up the number of sampled trajectories for ranking, highlighting a promising path toward \textit{test-time scaling}~\cite{deepseekai2025deepseekr1incentivizingreasoningcapability} for robot manipulation.
With the increase of $P$, the performance cosistently improves for larger $K$ values.
For smaller $K$ values, the performance initially improves and then reaches a plateau when $P=1000$.
This results indicate that data size and test-time computation should scale simultaneously.

\begin{figure}[!bp] 
    \centering
    \includegraphics[width=0.9\linewidth]{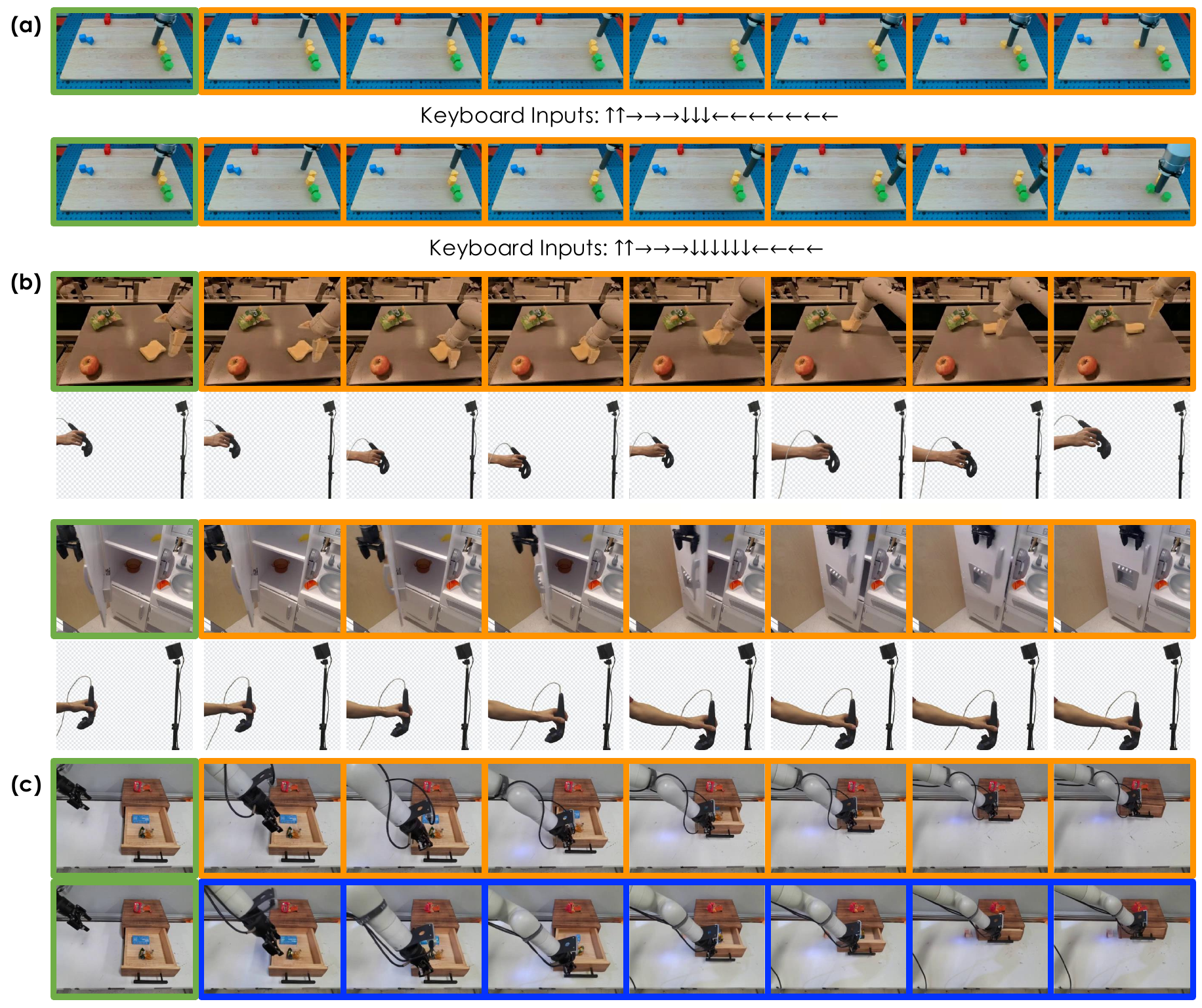}
    \caption{
       \textbf{Flexible Action Controllability.}
       We showcase controlling (a) the virtual robot in Language-Table with arrow keys on a keyboard and (b) the robot in RT-1 with a VR controller. 
       Predictions are in orange boxes; initial frames are in green boxes.
    }
    \label{fig:application}
\end{figure}

\paragraph{Real-Robot Experiments.}
We train \ourmethod on a real-robot dataset and perform experiments on three different tasks in the training dataset.
We leverage a goal-conditioned method which specifies the task via a goal image.
We use the similarity between the final image of the predicted video and the goal image as the value function for model-based planning.
We use a simple policy which samples 50 individual points from a sphere centered on the current end-effector position and generates a trajectory between the current position and each sampled point, resulting in $K=50$ different sampled trajectories.

Qualitative results are shown in Fig.~\ref{fig:mpc}. 
Quantitative results are shown in Tab.~\ref{tab:mpc}.
We experiment with two functions for similarity comparison: 1) mean squared error (MSE) and 2) cosine similarity of the feature extracted from ResNet50.
We observe that the MSE for value functions significantly outperformed the ResNet counterpart.
And both variants significantly outperform the policy without planning which randomly selects a trajectory for rollout.
These results demonstrate the potential of \ourmethod as a real-world manipulation world model for model-based planning.
More details and discussion can be found in Appendix~\ref{app:mpc}.

\subsection{Flexible Action Controllability}
\label{sec:control}
In this section, we perform qualitative experiments in which we ``control'' the virtual robot in two datasets, Language-Table~\cite{lynch2023interactive} and RT-1~\cite{brohan2022rt}, using trajectories collected with two distinct input sources: a keyboard and a VR controller.
Notably, the trajectories collected through these input sources exhibit distributions that deviate from those in the original dataset.
For Language-Table with a 2D translation action space, we use the arrow keys on the keyboard to input action trajectories.
For RT-1 with a 3D action space, we use a VR controller to collect action trajectories as input.
Specifically, we prompt \ourmethod with an image from each dataset and a trajectory collected with the keyboard or VR controller.
Fig.~\ref{fig:application} shows the video generated by \ourmethod.
\ourmethod is able to follow trajectories collected with different input sources and simulate robot-object interaction in a realistic and reasonable way. 
More importantly, it is able to robustly handle multimodality in generation.
Fig.~\ref{fig:application}(a) shows videos generated with an identical initial frame but different trajectories.
In Appendix~\ref{app:implausible}, we demonstrate that \ourmethod can also handle physically implausible trajectories robustly.

\section{Conclusion, Limitation and Future Work}
\label{sec:lim_con}
In this paper, we present \ourmethod, a novel 
world model that generates videos, with fine-grained robot-object interaction details, of a robot executing an action trajectory given historical observation.
We achieve precise alignments between actions and video frames via a novel frame-level action-conditioning module. 
Extensive experiments show the videos quality generated by \ourmethod is able to generate long-horizon and high-resolution videos that accurately simulate the robot trajectory rollouts. 
Additionally, we showcase that \ourmethod can be leveraged as a simulator for policy evaluation and a dynamics model for model-based planning to improve policy performance.
Similar to many other generative models, a limitation of \ourmethod is video generation is not real-time.
In the future, we plan to explore leveraging diffusion distillation~\citep{meng2023distillation} to accelerate generation speed.
In addition, we plan to investigate utilizing \ourmethod as a dynamics model and improve robot policies within the world model via reinforcement leanring~\cite{alonso2025diffusion}.



\paragraph{Acknowledgments.} We are grateful to Dr.~Zicong Hong for helping revise the paper and for the constructive feedback. 
This research was supported by fundings from the Hong Kong RGC General Research Fund (152169/22E, 152228/23E, 162161/24E), Research Impact Fund (No. R5060-19, No. R5011-23), Collaborative Research Fund (No. C1042-23GF), NSFC/RGC Collaborative Research Scheme (Grant No. 62461160332 \& CRS\_HKUST602/24), Areas of Excellence Scheme (AoE/E-601/22-R), and the InnoHK (HKGAI).

\clearpage
\bibliographystyle{unsrtnat}
\bibliography{main}

\clearpage

\beginappendix
\section{Additional Qualitative Results}
In this section, we present additional qualitative video results on the following:
1) Short Trajectories: We compare \ourmethod with baseline methods using short trajectories from RT-1, Bridge, and Language-Table. We also provide additional qualitative results of \ourmethod on RoboNet;
2) Long Trajectories: We compare \ourmethod with baseline methods in the long trajectories setting;
3) Scaling: We compare different sizes of \ourmethod;
4) Robustness to Physically Implausible Trajectories: We show that \ourmethod can handle physically implausible trajectories.

\subsection{Video Generation of Short Trajectories}
\label{app:short_compare}
\begin{figure*}[htbp] 
    \centering
    \includegraphics[width=0.9\linewidth]{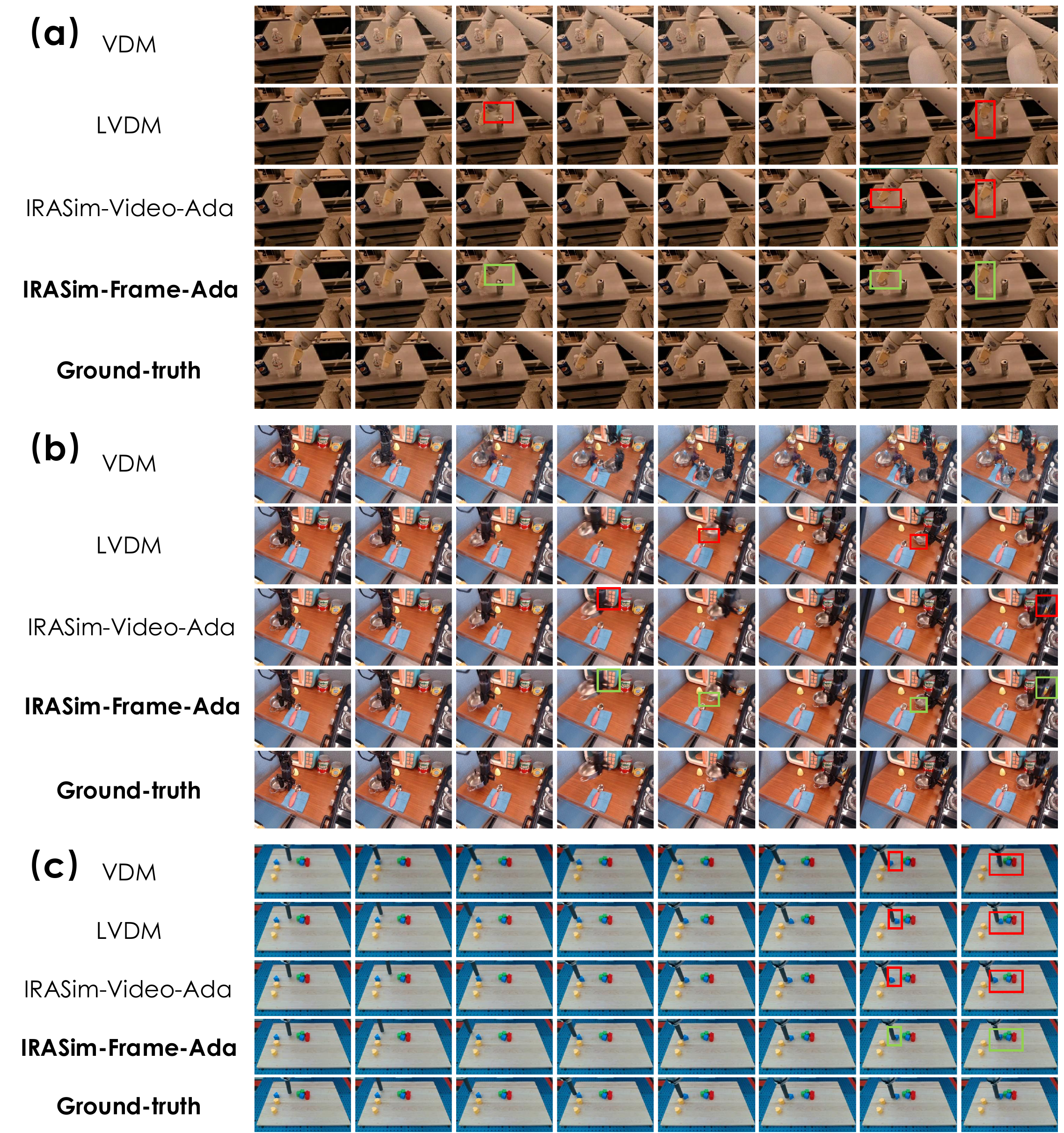}
    \caption{
        \textbf{Additional Qualitative Results on Video Generation of Short Trajectories.} We compare the results of different methods on (a) RT-1, (b) Bridge, and (c) Language-Table. Differences between \ourmethod-Frame-Ada and other methods are highlighted in green and red boxes.
    }
    \label{fig:short_compare}
\end{figure*}

Qualitative results are illustrated in 
Fig.~\ref{fig:short_compare} and Fig.~\ref{fig:robonet}. 
Fig.~\ref{fig:short_compare} demonstrates that \ourmethod-Frame-Ada surpasses other methods in aligning frames with actions and modeling the interaction between robots and objects. For the RoboNet dataset, we follow~\citet{wu2024ivideogpt} and use two frames as context for prediction. Fig.~\ref{fig:robonet} illustrates that \ourmethod is capable of simulating the manipulation of flexible objects, such as dragging clothes.

In terms of the number of context frames, we conduct an additional experiment on Bridge dataset and used 2 frames as context. The performance change is minor: the PNSR of using 1 context frame and 2 context frames are both 25. We hypothesize that the input trajectory itself contains sufficient information about velocity. Thus, including more context frames does not bring about significant improvement.

\subsection{Video Generation of Long Trajectories}
Qualitative results are illustrated in Fig.~\ref{fig:long_compare}.
\ourmethod-Frame-Ada generates consistent and long-horizon videos, accurately simulating the entire trajectory. 
Additionally, \ourmethod-Frame-Ada maintains its superior performance in frame-action alignment and robot-object interaction as observed in the short trajectory setting.

\subsection{Scaling}
\label{app:scale}
Qualitative results are shown in Fig.~\ref{fig:scale_compare}.
\ourmethod-Frame-Ada consistently improves the quality of the generated video in terms of reality and accuracy with the increase of model size.

\subsection{Robustness to Physically Implausible Trajectories}
\label{app:implausible}
We perform experiments on rolling out a physically implausible trajectory. In particular, we input a trajectory that commands the robot to move downward even after it touches the table. Physically, the robot cannot penetrate the table and thus will remain on the table even if the input control commands it to move down. We input this trajectory to \ourmethod to evaluate its performance in handling physically implausible trajectories. As shown in Fig.~\ref{fig:implausible}, \ourmethod can generate physically accurate videos where the robot stays on the table.

\begin{table*}[htbp]
    \centering
    \begin{tabular}{l|rr|rr|rr|rr}
        \toprule
        \textbf{Datasets} & \multicolumn{2}{c|}{\textbf{RT1}} & \multicolumn{2}{c|}{\textbf{Bridge}} & 
        \multicolumn{2}{c|}{\textbf{Language-Table}} & \multicolumn{2}{c}{\textbf{RoboNet}} \\
        \midrule
        \textbf{Data Split} & \textbf{Episode} & \textbf{Sample} & \textbf{Episode} & \textbf{Sample} & \textbf{Episode} & \textbf{Sample} & \textbf{Episode} & \textbf{Sample} \\
        \midrule
        Train & 82,069 & 2,314,893 & 25,460 & 482,701 & 170,256 & 1,483,133 & 162,161 & 2,540,500 \\
        Validation & 2,167 & 4,810 & 1,737 & 2,905 & 4,446 & 5,119 & - & - \\
        Test & 2,167 & 4,799 & 1,738 & 2,946 & 4,562 & 5,243 & 256 & 407 \\
        \bottomrule
    \end{tabular}
    \caption{Dataset Statistics. An "episode" is a single trial where the robot completes a task. A "sample" is a clip from an episode. 
    "-" indicates that we follow previous work and do not use a validation set.}
    \label{tab:dataset_summary}
\end{table*}

\section{Datasets}
\label{app:dataset}
\paragraph{Dataset Statistics.} We provide details on the four publicly available robot manipulation datasets: RT-1~\citep{brohan2022rt}, Bridge~\citep{walke2023bridgedata}, Language-Table~\citep{lynch2023interactive} and RoboNet~\citep{robonet}. 
A summary of the dataset statistics is presented in Table \ref{tab:dataset_summary}. 
For RT-1, Bridge and Language-Table, each training sample consists of a 4-second video clip containing 16 frames, extracted from an episode with a continuous sliding window. 
For testing and validation, frames are sampled at 16-frame intervals to reduce the number of evaluation videos and, consequently, lower evaluation costs. 
The original resolution for RT-1 is $256\times320$, for Bridge it is $480\times640$, and for Language-Table it is $360\times 640$. To ensure efficient training, we resize the Bridge videos to a resolution of $256\times320$ and the Language-Table videos to $288\times512$. 
For RoboNet, we follow \citet{wu2024ivideogpt} and use 2 frames as context to predict the next 10 frames at a resolution of $256 \times 256$.
Note that the mentioned "our own dataset" in Sec. 4.4 is similar in size to RT-1, and the action space is the same.

\paragraph{Action Space.} 
\label{app:action_space}
Different datasets have different action spaces.
In RT-1 and Bridge, a robot arm with a gripper moves in the 3D space to perform manipulation which interacts with objects in the scene.
The action spaces of RT-1 and Bridge consist of 1) 6-DoF arm actions in 3D space, $T \in {SE}(3)$, and 2) continuous gripper actions, $g \in [0, 1]$. 
In Language-Table, a robot arm moves in a 2D plane to move blocks with a cylindrical end-effector.
The action space of Language-Table is 2-DoF translation in 2D space, $p \in {R}^2$.
We convert the arm action of all datasets to relative delta actions.
Specifically, we specify the action of RT-1 and Bridge with a 7-dim vector, i.e., $a = [\Delta x, \Delta y, \Delta z, \Delta \alpha, \Delta \beta, \Delta \gamma, g]$ where $\Delta x$, $\Delta y$, and $\Delta z$ are the delta XYZ position; $\Delta \alpha$, $\Delta \beta$, and $\Delta \gamma$ are the delta Euler angles; $g$ indicates the gripper joint-angle position in the next step. For Language-Table, we specify the action with a 2-dim vector, i.e., $a = [\Delta x, \Delta y]$ which indicates the delta position in the xy-plane.
RoboNet is a large-scale robot manipulation dataset featuring 7 robot platforms with varying action spaces (2, 4, or 5 dimensions). Following~\citet{robonet}, to unify the data, a 5-dimensional vector is used to represent a universal action space, padding zeros for missing dimensions. This vector represents delta XYZ position, delta yaw angle, and gripper joint-angle value: $a = [\Delta x, \Delta y, \Delta z, \Delta \gamma, g]$. For instance, if a robot doesn't control the z-axis, $\Delta z$ is set to 0.

\section{\ourmethod Model Details}
In this section, we introduce more details about two types of trajectory condition methods in Sec. 3.3: \textit{Video-Level Condition} and \textit{Frame-Level Condition}.
\subsection{Video-Level Conditioning}
\label{app:model_details_video}
In video-level condition (Fig. 2(b)), we first obtain the conditioning embedding $\mathbf{c}_{ST}$ by adding the diffusion timestep embedding to the trajectory embedding. 
We then use $\mathbf{c}_{ST}$ to regress the scale parameters $\gamma$ and $\alpha$, as well as the shift parameters $\beta$. 
Specifically, the computation of the spatial block is as follows:
\begin{align}
    \label{eq:video_spatial_mha}
    \mathbf{x} &= \mathbf{x} + (1 + \alpha_1) \times \text{MHA}(\gamma_1 \times \text{LayerNorm}(\mathbf{x}) + \beta_1) \\
    \label{eq:video_spatial_ffn}
    \mathbf{x} &= \mathbf{x} + (1 + \alpha_2) \times \text{FFN}(\gamma_2 \times \text{LayerNorm}(\mathbf{x}) + \beta_2)
\end{align}
where $\mathbf{x}$, with a shape of $(N, P, D)$, denotes the token embeddings. 
$\mathbf{x}$ is reshaped as $(P, N, D)$ before entering the temporal block.
The computation of the temporal block is:
\begin{align}
    \label{eq:video_temporal_mha}
    \mathbf{x} &= \mathbf{x} + (1 + \alpha_3) \times \text{MHA}(\gamma_3 \times \text{LayerNorm}(\mathbf{x}) + \beta_3) \\
    \label{eq:video_temporal_ffn}
    \mathbf{x} &= \mathbf{x} + (1 + \alpha_4) \times \text{FFN}(\gamma_4 \times \text{LayerNorm}(\mathbf{x}) + \beta_4)
\end{align}
Note that layer normalization is performed before scaling and shifting.

\subsection{Frame-Level Condition}
\label{app:model_details_frame}
In frame-level condition (Fig. 2(c)), spatial attention blocks and temporal attention blocks are conditioned differently.
The derivation of the conditioning embedding for temporal attention blocks $\mathbf{c}_T$ is the same as in video-level condition, where we add the diffusion timestep embedding to the trajectory embedding.
Different frames are conditioned differently in spatial attention blocks.
We denote the conditioning embedding of spatial attention blocks for the i-th frame as $\mathbf{c}_S^{i}$.
To derive $\mathbf{c}_S^{i}$, the i-th action in the trajectory is first encoded to an embedding through a linear layer.
The diffusion timestep embedding is then added to the encoded embedding to obtain $\mathbf{c}_S^{i}$. 
We use $\mathbf{c}_S^1, \dots, \mathbf{c}_S^N$ and $\mathbf{c}_T$ to regress the corresponding scale parameters $\gamma$ and $\alpha$, as well as the shift parameters $\beta$.
While the computation of the temporal blocks is the same as the video-level condition~(Eq.~\ref{eq:video_temporal_mha} and~\ref{eq:video_temporal_ffn}), the computation of spatial blocks is different:
\begin{align}
\label{eq:frame_spatial_mha}
\mathbf{x}^i &= \mathbf{x}^i + (1+\alpha_1^i) \times \text{MHA}(\gamma_1^i \times \text{LayerNorm}(\mathbf{x}^i+\beta_1^i)), \\
\label{eq:frame_spatial_ffn}
\mathbf{x}^i &= \mathbf{x}^i + (1+\alpha_2^i) \times \text{FFN}(\gamma_2^i \times \text{LayerNorm}(\mathbf{x}^i+\beta_2^i)).
\end{align}
where $\alpha_1^i, \gamma_1^i, \beta_1^i, \alpha_2^i, \gamma_2^i, \beta_2^i$ denote the scale and shift parameters for the i-th frame.
They are regressed from $\mathbf{c}_{S}^i$.

\section{Baselines Details}
\label{app:baselines}
In this section, we detail the baseline implementation.
For VDM~\citep{ho2022video}, we leverage the implementation provided in \footnote{https://github.com/lucidrains/video-diffusion-pytorch}, which utilizes a 3D U-Net architecture for controllable video generation. 
We use only the model component from this code and keep the training setting consistent with \ourmethod. 
LVDM~\citep{he2023latent} employs the same model architecture as VDM.
It performs diffusion in the latent space while VDM performs diffusion in the pixel space. 
We use an MLP to encode the trajectory into an embedding.
It is then concatenated with the embedding of the diffusion timestep to form the conditioning embedding.
This is similar to the original methods in the paper where the text embedding is concatenated with the diffusion timestep embedding to form the conditioning embedding. 
The initial frame condition method of VDM and LVDM is the same as \ourmethod as described in Sec. 3.3. 
LVDM and \ourmethod share the same VAE model and training setting. 
Given that the resolution of Language-Table~\citep{lynch2023interactive} is up to $288\times512$, we resize the video to $144\times256$ in the training of VDM to make the computational cost affordable. 
During evaluation, we resize the generated video back to $288\times512$ for comparison with other methods. 
For RT-1 and Bridge, the training of VDM is performed at a resolution of $256\times320$. 
The training hyperparameters for VDM and LVDM are shown in Tab.~\ref{tab:vdm} and~\ref{tab:lvdm}. 
More training hyperparameters that share with \ourmethod can be found in Tab.~\ref{tab:irasim}.

We also briefly introduce the baseline details of iVideoGPT~\citep{wu2024ivideogpt} and MaskViT~\citep{maskvit}. Both of them use VQGAN~\citep{vqgan} as the image tokenizer and require additional finetuning it on RoboNet, while \ourmethod employs the VAE encoder from SDXL~\citep{podell2023sdxl} without the need for extra finetuning. Their parameter sizes are 436M and 228M, respectively. Moreover, iVideoGPT undergoes extensive pre-training on \citet{open_x_embodiment_rt_x_2023}, whereas \ourmethod achieves better video prediction performance with training only on RoboNet.

\begin{table*}[ht]
    \centering
    \begin{minipage}{0.5\linewidth}
        \centering
        
        \begin{tabular}{cc} 
        \toprule
        \textbf{Hyperparameter} & \textbf{Value} \\ \midrule
        Base channels & 64 \\
        Channel multipliers & 1,2,4,8 \\
        Num attention heads & 8 \\
        Attention head dimension & 32 \\
        Conditioning embedding dimension & 768 \\
        Input channels & 3 \\
        Parameters & 40M\\
        \bottomrule
        \end{tabular}
        \caption{Hyperparameters for VDM.}
        \label{tab:vdm}
    \end{minipage}%
    \hfill
    \begin{minipage}{0.5\linewidth}
        \centering
        
        \begin{tabular}{cc} 
        \toprule
        \textbf{Hyperparameter} & \textbf{Value} \\ \midrule
        Base channels & 288 \\
        Channel multipliers & 1,2,4,8 \\
        Num attention heads & 8 \\
        Attention head dimension & 32 \\
        Conditioning embedding dimension & 768 \\
        Input channels & 3 \\
        Parameters & 687M \\
        \bottomrule
        \end{tabular}
        \caption{Hyperparameters for LVDM.}
        \label{tab:lvdm}
    \end{minipage}
\end{table*}

\section{Training Details}
\label{app:training_details}
For all models, we use AdamW~\citep{adamw} for training.
We use a constant learning rate of 1e-4 and train for 300k steps with a batch size of 64. 
The gradient clipping is set to 0.1.
We found the training of \ourmethod very stable -- no loss spikes were observed even without gradient clipping.
However, loss spikes often occur in LVDM and VDM when gradient clipping is not used. 
Following~\citet{peebles2023scalable}, we utilize the Exponential Moving Average~(EMA) technique with a decay of 0.9999. 
All other hyperparameters are set the same as~\citet{peebles2023scalable}. Tab.~\ref{tab:irasim} lists further hyperparameters. 
All models are trained from scratch.
We utilize PNDM~\citep{liu2022pseudo} with 50 sampling steps for efficient video generation during evaluation. 
\ourmethod generates a 16-frame video with a duration of approximately 4 seconds, requiring only 30 seconds on an A100 GPU using 8GB of memory. 
Although there is still significant room for latency improvement, our method features high throughput and is memory-friendly during inference.

For scaling results in Fig. 4, the configurations of four different sizes of \ourmethod are shown in Tab.~\ref{tab:scale}. We study the scale performance of \ourmethod-Frame-Ada since it performs best.

The information about computing resources for training our \ourmethod is provided in Tab.~\ref{tab:compute_resources}.

\section{Evaluation Details}
\label{app:evaluate}
We introduce the evaluation details in this section. 
\paragraph{Evaluation Metrics.} Latent L2 loss and PSNR measure the L2 distance between the predicted video and the ground-truth video in the latent space and pixel space, respectively.
SSIM evaluates the similarity between videos in terms of image brightness, contrast, and structure.
FID and FVD assess video quality by analyzing the similarity of video feature distributions.

\paragraph{Evaluation Setup.} We evaluate the video quality generated by \ourmethod and the baselines under two settings: short trajectories and long trajectories. In the short trajectory setting, the input consists of one initial frame and a short trajectory containing 15 actions, resulting in the generation of 15 subsequent frames. These short trajectories are sampled from episodes using a sliding window with an interval of 16. In the long trajectory setting, the input comprises one initial frame and a complete long trajectory, with the output being the generated subsequent frames. The average lengths of the long trajectories are 42.5, 33.4, and 23.7 frames for RT-1, Bridge, and Language-Table, respectively. These lengths also represent the average number of frames for the generated long videos, which are produced in an autoregressive manner, as detailed in Sec. 4.1. The statistics of the generated short and long videos used for evaluation are presented in Tab.~\ref{tab:dataset_summary}.

\paragraph{Metric Calculation.} In all metric calculations, we ignore the initial frame and only evaluate the quality of the generated frames. For PSNR and SSIM, we refer to skimage~\footnote{https://scikit-image.org/docs/stable/api/skimage.metrics.html} for calculation. For FID and FVD, we refer to \footnote{https://github.com/mseitzer/pytorch-fid} and \footnote{https://github.com/universome/stylegan-v} for calculation, splitting the generated videos into frames and using their codebases to compute the FID and FVD values. However, we do not calculate FID and FVD metrics for long videos because we find that these metrics do not reflect human preferences well, even in the short trajectory setting. This could be because FID and FVD essentially calculate the similarity between the distributions of two datasets, whereas the \textit{trajectory-to-video} task is a reconstruction task, making reconstruction loss a more suitable evaluation metric.

\begin{table}[ht]
    \centering
    
        \begin{tabular}{cc} 
        \toprule
        \textbf{Hyperparameter} & \textbf{Value} \\ \midrule
        Layers & 28 \\
        Hidden size & 1152 \\
        Num attention heads & 16 \\
        Patch size & 2 \\
        Input channels & 4 \\
        Dropout & 0.1 \\
        Optimizer & AdamW($\beta=0.9,\beta=0.999$) \\
        Learning rate & 0.0001 \\
        Batch size & 64 \\
        Gradient clip & 0.1 \\
        Training steps & 3000000 \\
        EMA & 0.9999 \\
        Weight decay & 0.0 \\
        Prediction target & $\epsilon$ \\
        Parameters & 679M \\
        \bottomrule
        \end{tabular}
    \caption{Hyperparameters for training \ourmethod.}
    \label{tab:irasim}
\end{table}

\section{Real-Robot Model-based Planning Details}
\label{app:mpc}
In this section, we detail the real-robot model-based planning experiment. The experiment demonstrates that \ourmethod can effectively plan trajectories to finish manipulation tasks by generating the outcomes of executing different candidate trajectories.

\paragraph{Experiment Setup.} We follow ~\citet{fitvid} to set up this experiment. We implement a model-based policy to show the usefulness of \ourmethod.  Our policy consists of a sampling-based planner, a cost function, and \ourmethod as the dynamic function.  We first train \ourmethod with our own real robot dataset.  The input of our policy includes the initial image, the initial position of the end-effector, and a goal image to indicate the task.  The output is a predicted trajectory.  We use a simple sampling-based planner to generate candidate trajectories.  The planner samples 50 individual points from a circle centered on the initial end-effector position and then generates a trajectory between the initial position and each sampled point, resulting in 50 different candidate trajectories.  We input the initial image and each trajectory to \ourmethod to generate the video of executing each trajectory.  We use a cost function to calculate the similarity between each predicted video and the goal image. We experiment with 2 cost functions: 1) mean squared error (MSE) and 2) cosine similarity of the feature extracted from ResNet50. We execute the top 5 trajectories with the lowest cost (i.e., the predicted video most similar to the goal image) in the real world and calculate the average success rate.  The experiment is repeated three times for each task. 

\paragraph{Results} Qualitative results are shown in Fig. 7. Quantitative results are shown in Tab. 5. We compare our method with a baseline that randomly picks a trajectory from the 50 candidates. The results show that using \ourmethod significantly increases the success rate compared to the random baseline.

\paragraph{Discussion About Cost Function.} We also explore how different cost functions impact the model's performance. 
We find that the MSE cost function is generally superior to the ResNet cost function. 
But the MSE cost function is not always perfect; sometimes it selects incorrect prediction videos, leading to task failure.
This suggests that we need to explore better cost functions in future work, considering that the success rate is influenced by both the accuracy of video prediction and the accuracy of the cost function. A suboptimal cost function could affect the evaluation of the video prediction model, as also mentioned by iVideoGPT~\citep{wu2024ivideogpt} and VLMPC~\citep{zhao2024vlmpcvisionlanguagemodelpredictive}. 

\paragraph{Discussion About Sample Policy.} Although we use a simple sampling-based planner as the sample policy in this experiment, we note that \ourmethod can be combined with any policy that has trajectory sampling capabilities~(i.e., action chunk techniques~\citep{chi2023diffusionpolicy, act}). The performance and range of tasks that \ourmethod can handle could be further enhanced by adopting a more advanced policy~\citep{chi2023diffusionpolicy, act}, which is capable of generating more precise and complex trajectories.

\section{Human Preference Evaluation}
\label{app:human}
Five participants took part in the human evaluation.
For each participant, we randomly sampled 10 ground-truth video clips from the test set for each of the 3 datasets.
And for each video clip, we juxtapose the predictions of \ourmethod-Frame-Ada with those of VDM, LVDM, and \ourmethod-Video-Ada (Fig.~\ref{fig:human_screenshot}).
Thus, a participant evaluated $90$ pairs of video clips.
Note that the orders of the juxtaposition are random for different clips.
See the caption of Fig.~\ref{fig:human_screenshot} for more details. 
We compare the results of all evaluated video clips and calculate the win, tie, and loss rates.
The screenshot of the GUI used in the human evaluation is shown in Fig.~\ref{fig:human_screenshot}.
The full text of the instruction given to participants is as follows:

\begin{tcolorbox}[colback=gray!10!white, colframe=gray!50!black, title=Evaluation Instructions]
You are asked to choose the more realistic and accurate video from two generated videos (shown above).
The ground-truth video is given as a reference (shown below).
Please carefully examine the given videos.
If you can find a significant difference between the two generated videos, you may choose which one is better immediately.
If not, please replay the videos more times.
If you are still not able to find differences, you may choose the "similar" option.
Please do not guess.
Your decision needs solid evidence.

\end{tcolorbox}

\begin{table*}[ht]
\centering
\begin{tabular}{ccccc}
\toprule
Model & Layers & Hidden size & Num attention heads & Parameters \\ \midrule
\ourmethod-S & 12 & 384 & 6 & 33M \\
\ourmethod-B & 12 & 768 & 12 & 132M \\ 
\ourmethod-L & 24 & 1024 & 16 & 461M \\ 
\ourmethod-XL & 28 & 1152 & 16 & 679M \\ \bottomrule
\end{tabular}
\caption{Model Sizes. We use \ourmethod as an abbreviation of \ourmethod-Frame-Ada for brevity.}
\label{tab:scale}
\end{table*}

\begin{table*}[ht]
\centering
\begin{tabular}{cccc}
\toprule
Dataset & Concurrent GPUs & GPU Hours &  GPU type \\ \midrule
RT-1 & 32 & 2381 & A800~(40 GB) \\
Bridge & 32 & 2371 & A800~(40 GB) \\ 
Lanaguge-Table & 32 & 2369 & A100~(80 GB) \\ \bottomrule
\end{tabular}
\caption{Compution resources for training \ourmethod.}
\label{tab:compute_resources}
\end{table*}

\begin{figure*}[htbp] 
    \centering
    \includegraphics[width=0.9\linewidth]{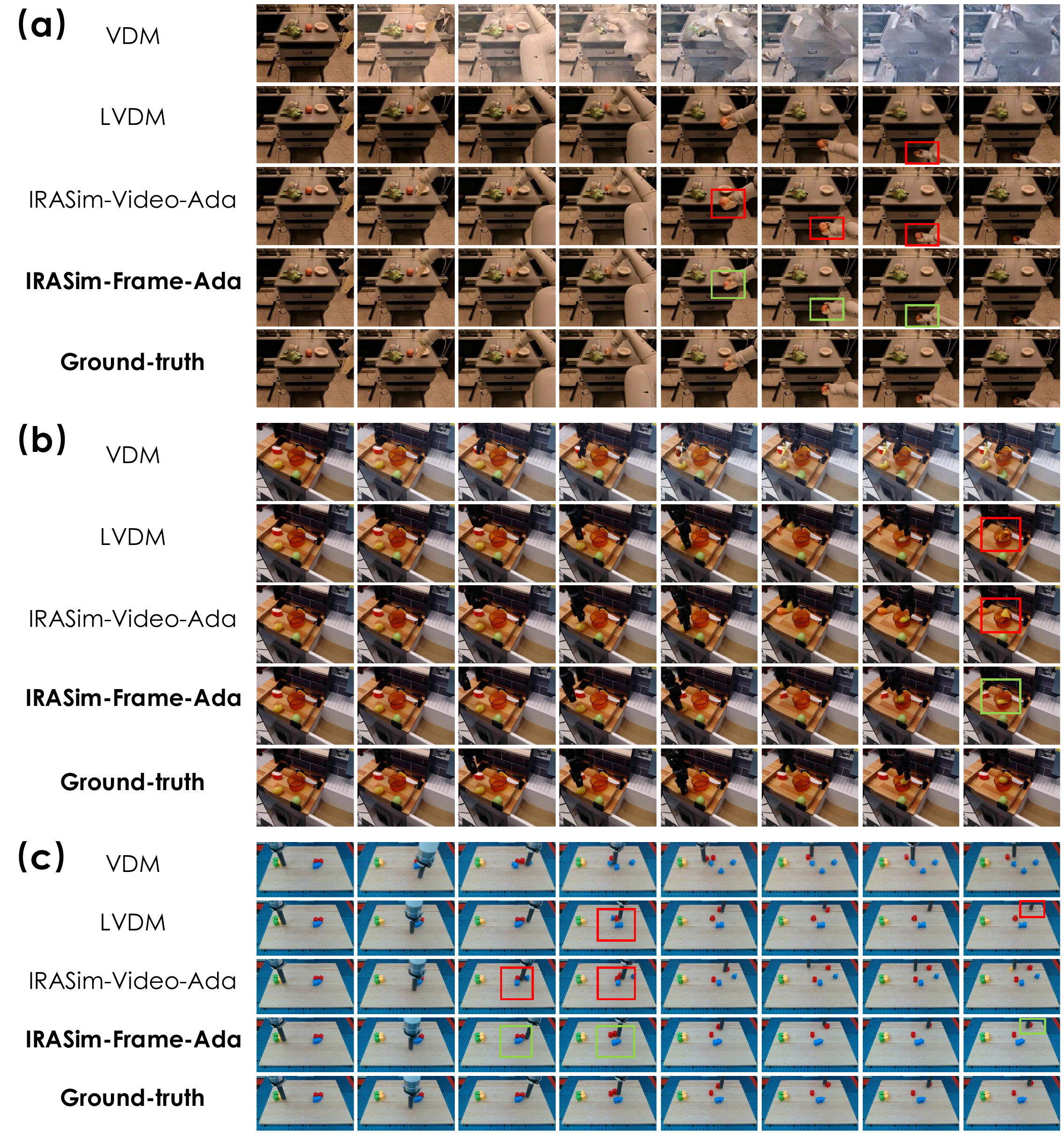}
    \caption{
        \textbf{Additional Qualitative Results on Video Generation of Long Trajectories.} 
        We compare the results of different methods on (a) RT-1, (b) Bridge, and (c) Language-Table. 
        Differences between \ourmethod-Frame-Ada and other methods are highlighted in green and red boxes.
        Note that the input trajectory is the entire trajectory of an episode.
    }
    \label{fig:long_compare}
\end{figure*}

\begin{figure*}[htbp] 
    \centering
    \includegraphics[width=0.9\linewidth]{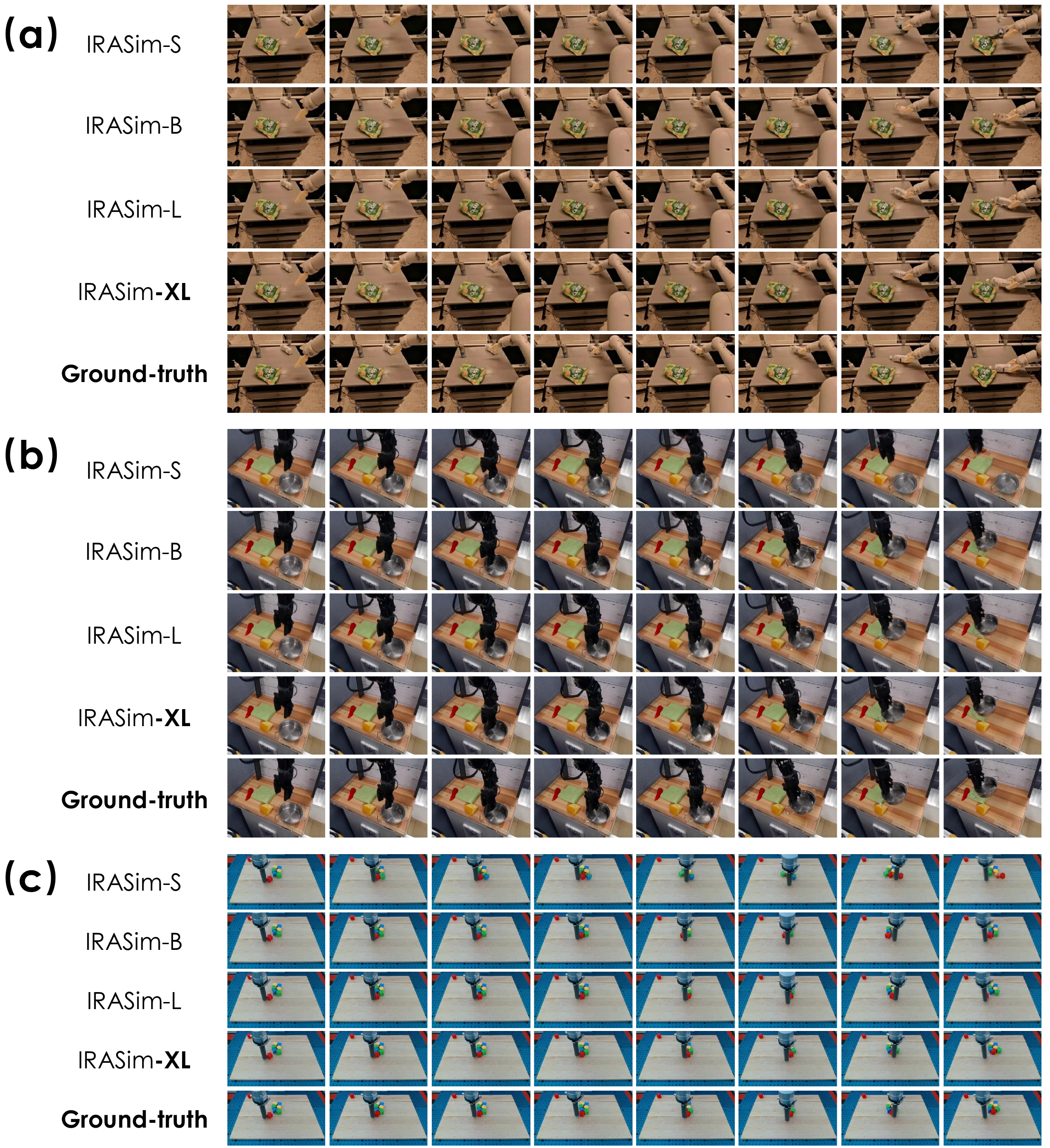}
    \caption{
        \textbf{Additional Qualitative Results on Scaling.}
        We compare the results of \ourmethod-Frame-Ada with different model sizes on (a) RT-1, (b) Bridge, and (c) Language-Table.
    }
    \label{fig:scale_compare}
\end{figure*}

\begin{figure*}[htbp] 
    \centering
    \includegraphics[width=1.0\linewidth]{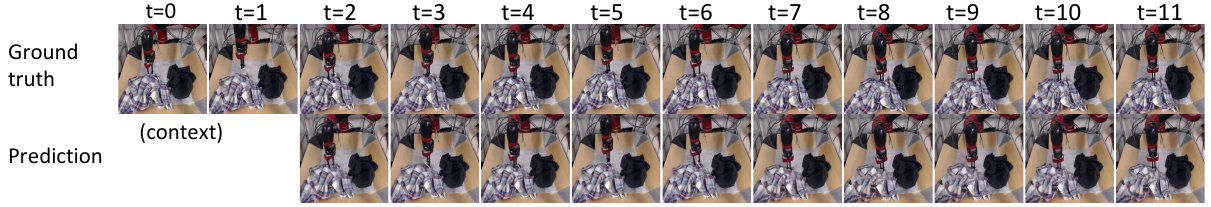}
    \caption{
     Quantitative results of \ourmethod-Frame-Ada on the RoboNet dataset. The robot is dragging the clothes, indicating that \ourmethod is capable of simulating the deformation of flexible objects.
    }
    \label{fig:robonet}
\end{figure*}

\begin{figure*}[htbp] 
    \centering
    \includegraphics[width=1.0\linewidth]{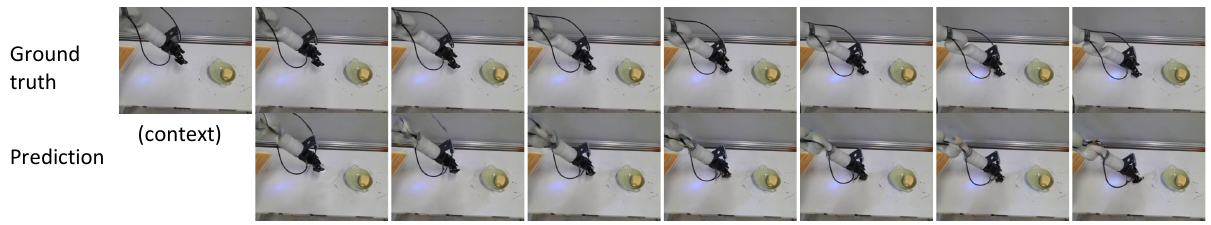}
    \caption{
    Quantitative results show that \ourmethod is robust to physically implausible trajectories. We control the robot to poke at the table and record the command trajectory, which is very dangerous as it could damage the robot. As a result, the robotic arm is blocked by the table. We find that executing the same trajectory in \ourmethod yields similar results, rather than the robotic arm passing through the table. This indicates that \ourmethod has a certain understanding of the physical laws of the real world.
    }
    \label{fig:implausible}
\end{figure*}

\begin{figure*}[htbp] 
    \centering
    \includegraphics[width=1.0\linewidth]{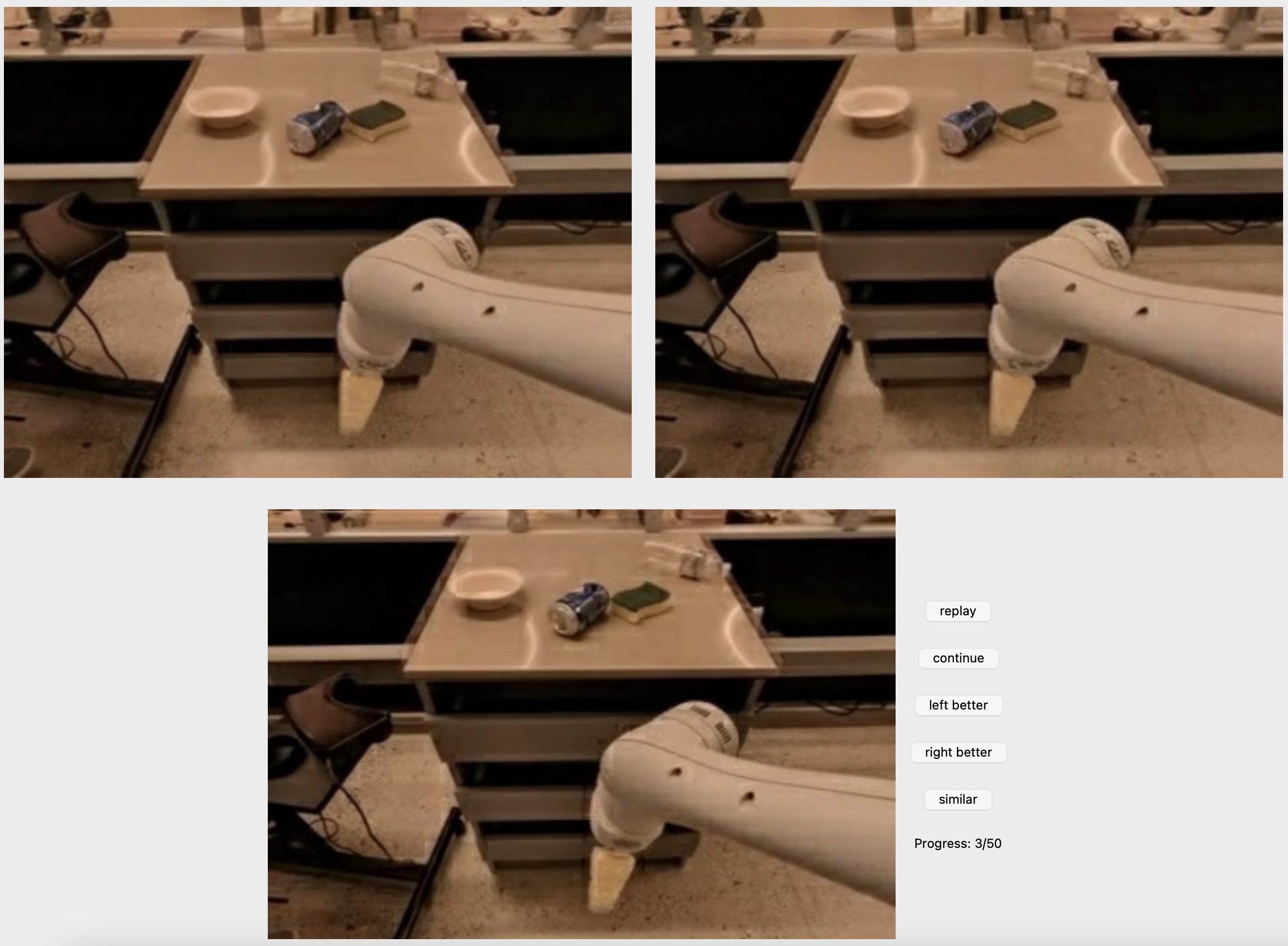}
    \caption{
    \textbf{Screenshot of the GUI in Human Preference Evaluation.}
    The two videos in the upper row are generated by \ourmethod-Frame-Ada and a comparing method, arranged in a \textbf{random} left-right order. The video in the lower row is the ground-truth video.
    }
    \label{fig:human_screenshot}
\end{figure*}

\end{document}